\documentclass[11pt]{article}

\usepackage{hydrosophy}

\usepackage{mathtools}

\usepackage{subcaption}
\usepackage{wrapfig}
\usepackage{tabularx}
\usepackage{multirow}
\usepackage{adjustbox}
\usepackage{colortbl}
\definecolor{headergray}{gray}{0.90}
\definecolor{lavender}{RGB}{230,230,250}

\usepackage{pifont}
\usepackage{marvosym}
\usepackage{accsupp}
\newcommand{\cmark}{\textcolor{green!60!black}{\ding{51}}}
\newcommand{\xmark}{\textcolor{red!75!black}{\ding{55}}}
\newcommand{\meanstd}[2]{\begin{tabular}[t]{@{}c@{}}#1\\[-2pt]{\scriptsize\color{hsgray}$\pm$#2}\end{tabular}}
\newcommand{\corrmark}{\BeginAccSupp{method=hex,unicode,ActualText=2709}\Letter\EndAccSupp{}}

\usepackage{xurl}

\usepackage{algorithm}
\usepackage{algorithmic}
\newcommand{\INPUT}{\STATE \textbf{Input:} }
\newcommand{\OUTPUT}{\STATE \textbf{Output:} }

\newcommand{\githubicon}{\raisebox{-1.5pt}{\includegraphics[height=1.05em]{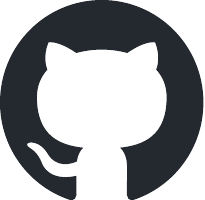}}}
\newcommand{\hficon}{\raisebox{-1.5pt}{\includegraphics[height=1.05em]{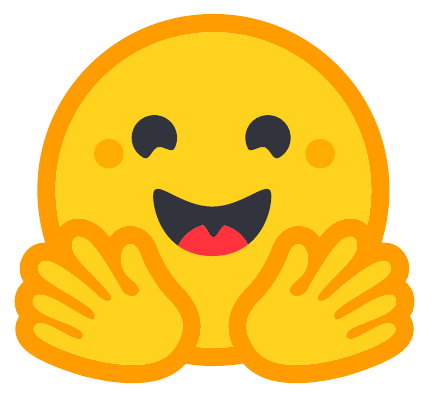}}}

\title{Breaking Bad Molecules: Are MLLMs Ready for Structure-Level Molecular Detoxification?}
\shorttitle{ToxiMol}

\author[1,2,3,*]{Fei Lin}
\author[1,3,*]{Ziyang Gong}
\author[4,*]{Cong Wang}
\author[1,2]{Tengchao Zhang}
\author[4]{Yonglin Tian}
\author[2]{Yining Jiang}
\author[5]{Ji Dai}
\author[4]{Chao Guo}
\author[6]{Xiaotong Yu}
\author[3,\corrmark]{Xue Yang}
\author[7]{Gen Luo}
\author[2,4,\corrmark]{Fei-Yue Wang}

\affiliation[1]{HydroSophy}
\affiliation[2]{Macau University of Science and Technology}
\affiliation[3]{Shanghai Jiao Tong University}
\affiliation[4]{Institute of Automation, Chinese Academy of Sciences}
\affiliation[5]{Ningbo University}
\affiliation[6]{Institute of Process Engineering, Chinese Academy of Sciences}
\affiliation[7]{Shanghai Artificial Intelligence Laboratory}

\authornote[*]{Equal contribution.}
\authornote[\corrmark]{Corresponding authors.}

\metadata[\githubicon\ Code]{\url{https://github.com/HydroSophy/ToxiMol}}
\metadata[\hficon\ Dataset]{\url{https://huggingface.co/datasets/HydroSophy/ToxiMol-benchmark}}

\patchcmd{\maketitle}{\vskip 24pt}{\vskip 12pt}{}{}
\patchcmd{\maketitle}{\vskip 15pt}{\vskip 10pt}{}{}
\patchcmd{\abstract}{before skip=12pt,after skip=18pt}{before skip=6pt,after skip=8pt,top=6pt,bottom=6pt}{}{}

\begin{document}

\maketitle

\begin{abstract}
Toxicity remains a leading cause of early-stage drug development failure. Despite advances in molecular design and property prediction, the task of \textit{molecular toxicity repair}—generating structurally valid molecular alternatives with reduced toxicity—has not yet been systematically defined or benchmarked. To fill this gap, we introduce \textbf{ToxiMol}, the first benchmark task for general-purpose Multimodal Large Language Models (MLLMs) focused on molecular toxicity repair. We construct a standardized dataset covering 11 primary tasks and 660 representative toxic molecules spanning diverse mechanisms and granularities. We design a prompt annotation pipeline with mechanism-aware and task-adaptive capabilities, informed by expert toxicological knowledge. In parallel, we propose an automated evaluation framework, \textbf{ToxiEval}, which integrates toxicity endpoint prediction, synthetic accessibility, drug-likeness, and structural similarity into a high-throughput evaluation chain for repair success. We systematically assess 43 mainstream general-purpose MLLMs and conduct multiple ablation studies to analyze key issues, including evaluation metrics, candidate diversity, and failure attribution. Experimental results show that although current MLLMs still face significant challenges on this task, they begin to demonstrate promising capabilities in toxicity understanding, semantic constraint adherence, and structure-aware editing.
\end{abstract}

\section{Introduction}

\begin{wrapfigure}{r}{0.48\textwidth}
  \vspace{-36pt}
  \captionsetup{font=scriptsize,skip=3pt}
  \centering
  \includegraphics[width=0.94\linewidth]{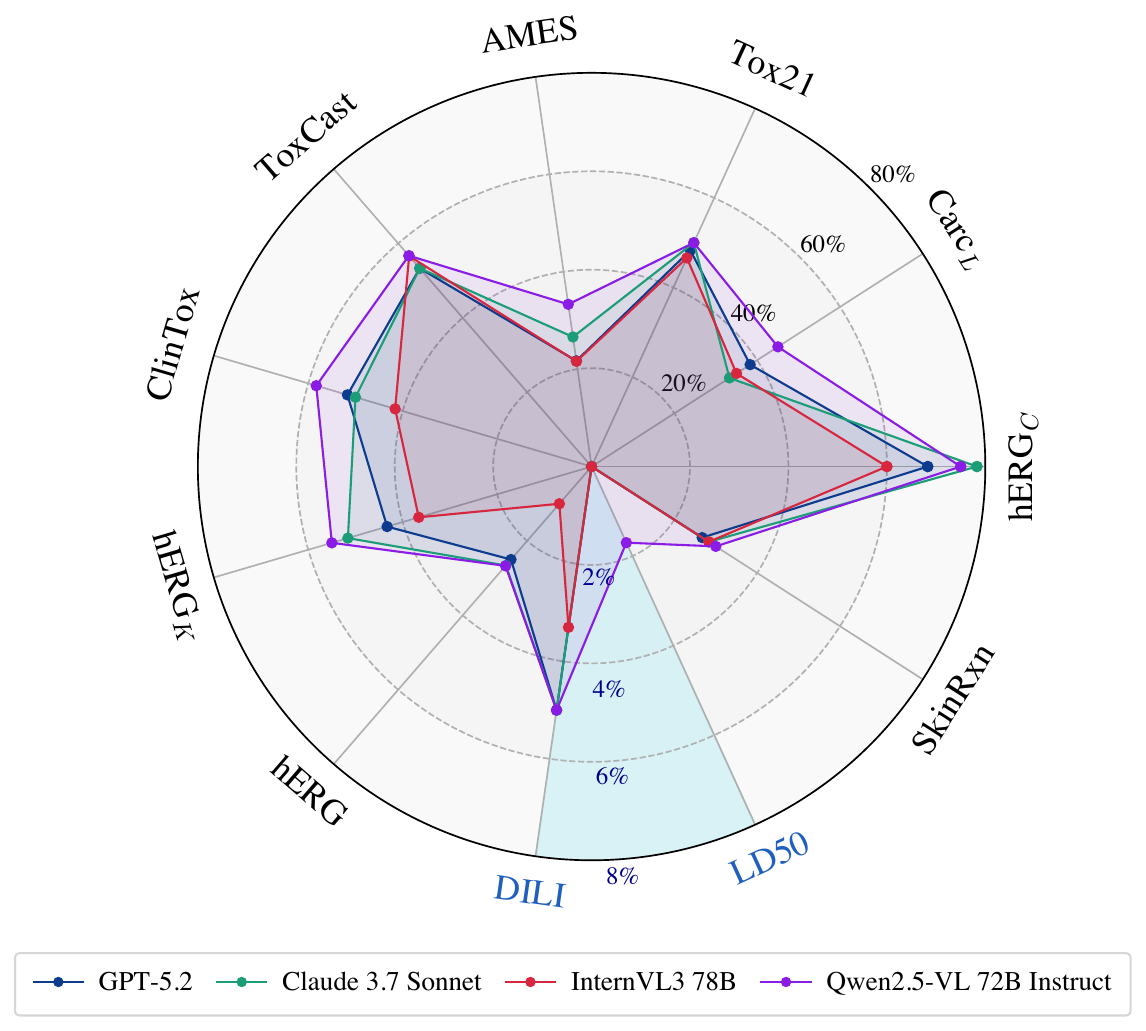}
  \caption{Success rates (\%) of representative MLLMs~\cite{openai_gpt52_2025, claude37sonnet2024, wang2025internvl3, bai2025qwen2} on 11 toxicity repair tasks. Highlighted tasks use a 0--8\% scale.}
  \label{fig:radar}
  \vspace{-8pt}
\end{wrapfigure}

In drug discovery, approximately 90\% of candidate compounds fail due to poor Absorption, Distribution, Metabolism, Excretion, and Toxicity (ADMET) properties, with toxicity issues being among the primary causes of attrition during preclinical development~\cite{sun202290, li2004accurate, van2019limitations}. Traditional toxicity mitigation strategies rely on structural modifications, scaffold replacement, or fragment-based optimization for small-molecule drugs~\cite{nassar2004improvingthe, mullins2017toxicity, dey2008fragment}. These processes are not only heavily dependent on expert knowledge but also require extensive iterative experimentation, resulting in high complexity, long development cycles, and significant costs~\cite{basile2019artificial}. In recent years, the rapid advancement of multimodal fusion and Large Language Models (LLMs) has led to the emergence of general-purpose Multimodal Large Language Models (MLLMs~\cite{openai2024gpt4v, luo2024mono, InternVL3.0-8B, bai2023qwenvl}) with strong capabilities in cross-modal perception and reasoning~\cite{kuang2025natural}. These MLLMs are increasingly explored in highly structured scientific domains such as chemistry and life sciences~\cite{ramos2025review, lin2025autonomous}. Against this backdrop, a critical question arises: do general-purpose MLLMs possess the capacity to recognize and repair toxic molecules, and can they effectively support the ``detoxification'' objectives of molecular design?

Molecular 2D images have been widely used in a variety of downstream tasks, including scientific literature mining, molecular structure editing, and molecular property prediction, among others~\cite{wang2025image, zeng2022accurate, campos2021img2smi}. Recent studies have begun to explore the application of general-purpose MLLMs to tasks such as molecule grounding, toxicity prediction, and structure editing, achieving promising progress in multiple scenarios~\cite{wu2025molground, yang2025large, li2024geometry, jang2024chain, rosenwasser2025leveraging, lv2025exploiting}. However, the task of molecular toxicity repair—a critical capability in drug design—has yet to be systematically defined, leaving a significant gap in current research. To address this, we introduce \textbf{ToxiMol}, the first comprehensive benchmark for molecular toxicity repair tailored to general-purpose MLLMs. This task extends beyond conventional domains such as toxicity prediction, ADMET modeling, and property-driven optimization~\cite{niu2024pharmabench, aksamit2024hybrid, fallani2025pretraining}, aiming to rigorously assess the ability of MLLMs to perform complex molecular structure refinement. Specifically, the task requires the model to take as input a toxic molecule represented by its Simplified Molecular Input Line Entry System (SMILES), 2D structural image, and a natural language description of the repair objective, then identify potential toxicity endpoints, interpret semantic constraints, and generate structurally similar substitute molecules that successfully eliminate toxic fragments while satisfying drug-likeness and synthetic feasibility requirements. This process poses multifaceted challenges to the model, including toxicological knowledge representation, fine-grained perception of molecular structure, semantic comprehension of complex instructions, and high-reliability response generation.

We construct a standardized molecular toxicity repair dataset by reselecting and integrating high-quality annotated samples from multiple toxicity prediction tasks on the Therapeutics Data Commons (TDC) platform~\cite{velez2024tdc, huang2021therapeutics, huang2022artificial} to systematically evaluate this highly challenging and practically valuable task. The dataset covers 11 primary tasks, with two of them containing 12 and 10 sub-tasks, respectively, and includes a total of 660 representative toxic molecules spanning diverse mechanisms and varying toxicity granularities. For each task, we design prompt templates with input from \textbf{human toxicology experts} and construct a mechanism-aware, task-adaptive prompt annotation pipeline. To enable automated, multidimensional, and objective evaluation of the generated molecules, we further propose an evaluation framework—\textbf{ToxiEval}—that integrates toxicity endpoint prediction, synthetic accessibility, drug-likeness scores, and structural similarity into a high-throughput molecular assessment chain to determine the success of toxicity repair.

We conducted a systematic evaluation of 43 mainstream general-purpose MLLMs, including both closed-source and open-source models, and designed multiple ablation studies to analyze various aspects of the task. These include analysis of structural validity, strategies for combining multi-dimensional criteria, the influence of candidate set size, and failure mode attribution. Experimental results show that while the overall success rate of current general-purpose MLLMs on the toxicity repair task remains relatively low, the models are beginning to demonstrate initial potential for addressing this challenging problem. In summary, the main contributions of this work are as follows:

\begin{itemize}[leftmargin=15pt]

\item \textbf{A novel task and challenge for general-purpose MLLMs.} We introduce \textbf{ToxiMol}, the first benchmark for \textit{molecular toxicity repair}, as a new standard to evaluate the generalization and execution capabilities of general-purpose MLLMs on highly complex scientific problems.

\item \textbf{A standardized toxicity repair dataset and automated evaluation framework.} We construct a benchmark dataset featuring diverse mechanisms and controllable granularity and propose \textbf{ToxiEval}, an automated, multi-dimensional evaluation framework that enables fair, transparent, and quantitative comparison of model performance under complex repair objectives.

\item \textbf{Extensive evaluation and systematic analysis.} We conduct a comprehensive assessment of 43 mainstream general-purpose MLLMs across multiple tasks, models, and evaluation criteria, and perform in-depth ablation studies to examine the effects of various factors on repair success, revealing both the potential and current limitations of MLLMs in toxicity repair.

\end{itemize}

\section{Related Work}

\subsection{Molecular Toxicity Repair}

The molecular toxicity repair task lies at the intersection of two major research directions: molecular toxicity prediction and molecular structure editing. Traditional toxicity prediction approaches rely on rule-based expert systems, Quantitative Structure-Activity Relationship (QSAR) models~\cite{greene2002computer, gatnik2010review, benigni2008predictivity}, and Graph Neural Networks (GNNs)~\cite{monem2025drug, bai2025machine}. More recently, the rise of LLMs has introduced a new paradigm, with zero-shot models such as Tx-LLM~\cite{chaves2024tx}, MolE~\cite{mendez2024mole}, and TxGemma~\cite{wang2025txgemma} demonstrating strong cross-task generalization capabilities and significantly expanding the expressive framework of toxicity prediction. In contrast, molecular structure editing focuses on property-guided reconstruction of molecules. Representative approaches include GNN-based property-driven editing~\cite{bongini2021molecular, zhang2024deep}, molecular diffusion models~\cite{wang2024diffusion, schneuing2024structure}, and LLM-based structure generation~\cite{liu2023multi, liu2024conversational, li2024geometry, averly2025liddia, ye2025drugassist}. These methods typically optimize for a single target property—such as solubility, permeability, or drug-likeness—by generating new structures that meet specific~constraints.

In contrast, molecular toxicity repair remains in its early exploratory phase, treated as a standalone task and \textbf{yet to be systematically defined}. Traditional methods primarily rely on rule-based toxicophore databases such as ToxAlerts~\cite{sushko2012toxalerts}, which lack the capability for structure generation or automated repair~\cite{ehmki2019comparing, schmidt2019comparing}. Recent studies have attempted to incorporate toxicity metrics into joint ADMET optimization frameworks~\cite{du2023admet}. Some works employ deep neural network models, such as Transformers, for ADMET property prediction and regression modeling~\cite{monem2025drug, yang2023transformer, fallani2025pretraining, bulusu2016modelling, zitnik2018modeling}, while others have begun to explore LLM-based multitask benchmarks for ADMET tasks~\cite{niu2024pharmabench, ye2025drugassist, abunasser2024large}. However, there remains a lack of systematic approaches for modeling and evaluating the specific goal of “toxicity repair,” i.e., generating structurally valid, low-toxicity alternatives from toxic molecules.

\subsection{QA Benchmarks for MLLMs}


For evaluating general purpose MLLMs, researchers have developed systematic benchmarks across various task dimensions, including image captioning, Visual Question Answering (VQA), and code generation~\cite{kil2024mllm, kuang2025natural, wu2024plot2code}. In the chemistry and pharmaceutical domains, emerging benchmarks such as Molecule Captioning~\cite{li2024towards, ha2025mv}, MolTextQA~\cite{laghuvarapu2024moltextqa}, Mol-Hallu~\cite{li2025detect}, $\mathrm{S}^2$-Bench~\cite{li2025speaktostructure}, and MolPuzzle~\cite{guo2024can} have been proposed to assess model capabilities in molecular structure understanding, textual description, and task reasoning~\cite{huang2024chemeval, li2025chemvlm}. However, no established benchmark remains for evaluating MLLMs in the critical setting of toxicity repair. To our knowledge, this work is the \textbf{first to introduce toxicity repair as a benchmark task for general-purpose MLLMs} and to systematically evaluate their performance in molecular structure optimization and image-based chemical analysis.

\section{ToxiMol Benchmark}
\label{sec:ToxMol}

\subsection{Overview}
\label{sec:Overview}

\paragraph{Task Definition.} The ToxiMol Benchmark defines a toxicity repair task as follows: Given a toxicity-positive small molecule represented by its SMILES string $\text{SMILES}_i^{\text{raw}}$, its corresponding 2D molecular structure image $\text{IMG}_i$ (rendered via RDKit~\cite{rdkit} following international chemical depiction conventions, e.g., oxygen atoms in red, nitrogen in blue, with a unified and standardized rendering configuration to mitigate the influence of image style variations on the model), and a repair prompt $\mathcal{P}_i$, these inputs are fed into a MLLM denoted as $\mathcal{M}$. The model is instructed to generate a set of \textbf{3 candidate repaired molecules}, denoted by $\mathcal{S}_i^{\text{rep}}:= \{ \text{SMILES}_i^{\text{rep}(1)},\ \text{SMILES}_i^{\text{rep}(2)},\ \text{SMILES}_i^{\text{rep}(3)} \}$. If any of the candidate molecules in $\mathcal{S}_i^{\text{rep}}$ satisfies the multi-criteria evaluation defined by the ToxiEval framework, the repair is considered successful; otherwise, it is deemed a failure. The formal definition of this task is presented below.
\[
(\text{SMILES}_{i}^{\mathrm{raw}},\ \text{IMG}_{i},\ \mathcal{P}_{i})
\;\xrightarrow{\mathcal{M}}\;
\mathcal{S}_{i}^{\mathrm{rep}}
\;\Rightarrow\;
\operatorname{Success}(i)=
\begin{cases}
1, & \exists\, s\in\mathcal{S}_{i}^{\mathrm{rep}}\ \text{s.t.}\ \operatorname{ToxiEval}(s)=1 \\[2pt]
0, & \text{otherwise}
\end{cases}
\]

\paragraph{Data Statistics.} To construct a representative and challenging benchmark for molecular toxicity repair, we build upon all toxicity prediction tasks under the ``Single-instance Prediction Problem'' category in the TDC platform~\cite{velez2024tdc, huang2021therapeutics, huang2022artificial}, and systematically define $11$ primary toxicity repair tasks. All task names are kept consistent with the original datasets, namely: AMES~\cite{xu2012silico}, hERG~\cite{wang2016admet}, hERG\_Karim~\cite{karim2021cardiotox}, hERG\_Central~\cite{du2011hergcentral}, Tox21~\cite{huang2016tox21challenge}, ToxCast~\cite{richard2016toxcast}, Carcinogens~\cite{lagunin2009computer,cheng2012admetsar}, SkinReaction~\cite{alves2015predicting}, DILI~\cite{xu2015deep}, LD50~\cite{zhu2009quantitative}, and ClinTox~\cite{gayvert2016data}. Among them, Tox21 retains all $12$ subtasks, whereas ToxCast randomly selects $10$. These tasks cover approximately 30 small-molecule toxicity mechanisms, with detailed descriptions provided in Appendix~\ref{toxicity_mechanisms}.

During sample construction, for all tasks except \texttt{LD50}, we select all toxic molecules with label $Y=1$ from the original datasets; for the \texttt{LD50} task, we select all highly toxic molecules satisfying $\log(\text{LD}_{50}) < 2$. Since the number of available toxic molecules varies substantially across tasks (ranging from $60$ to $13{,}744$), with the \texttt{Carcinogens} task containing only $60$ toxic samples and thus forming the bottleneck for sampling scale, we adopt a task-balanced sampling strategy to ensure fairness and comparability across tasks and to prevent tasks with larger sample sizes from dominating the overall evaluation. Specifically, the sampling size for all tasks is uniformly set to $n = 60$. For standard tasks, $60$ toxic molecules are sampled per task; for the \texttt{Tox21} and \texttt{ToxCast} tasks, a stratified balanced sampling strategy is adopted: for \texttt{Tox21}, $5$ molecules are sampled for each of the $12$ subtasks ($12 \times 5 = 60$); for \texttt{ToxCast}, $6$ molecules are sampled for each of the $10$ subtasks ($10 \times 6 = 60$). This design ensures that each primary task contributes equally to the final evaluation metrics.

\begin{figure}[!htbp]
  \centering
  \includegraphics[width=0.6\linewidth]{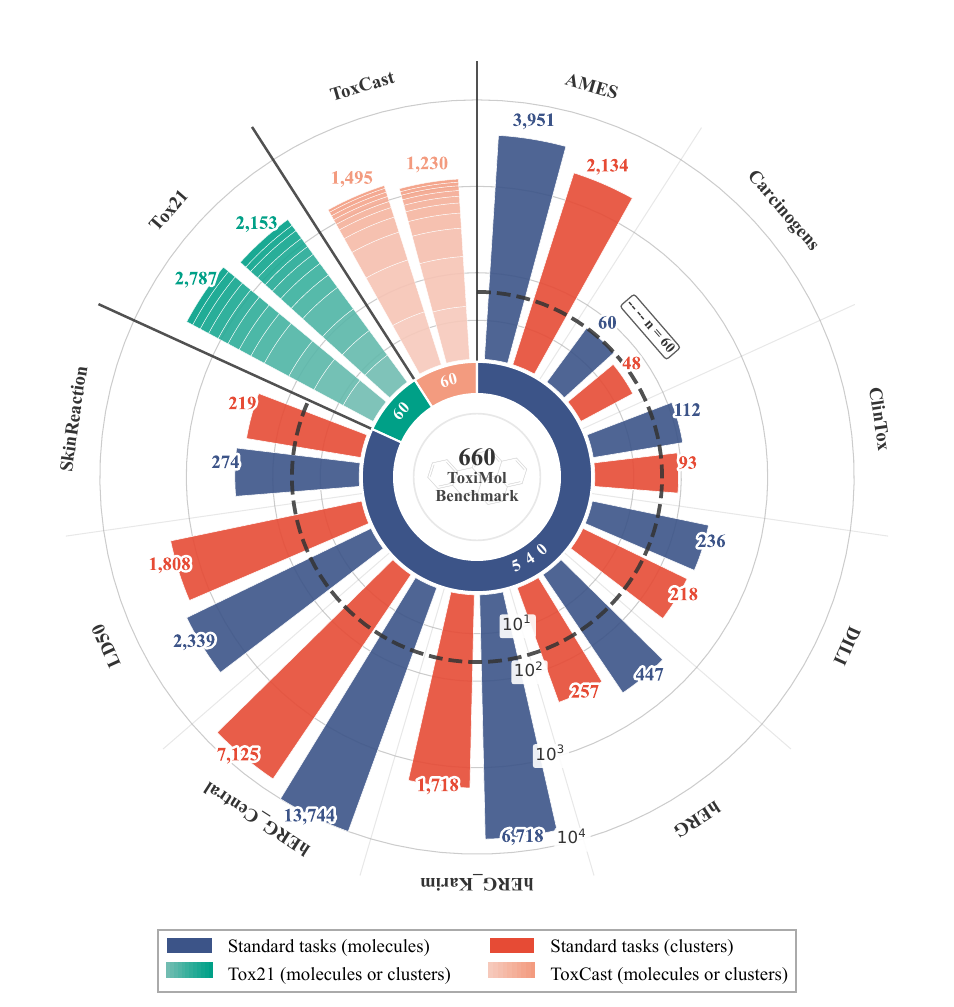}
  \caption{Statistics of toxic molecules and the number of clusters for each task. Values are plotted along the radius on a logarithmic scale ($10^{1}$--$10^{4}$); the dashed ring at $n=60$ indicates the balanced sampling size determined by the Carcinogens bottleneck constraint.}
  \label{fig:dataset}
\end{figure}

For sampling methodology, we adopt a structure-aware representative sampling strategy: ECFP4~\cite{rogers2010extended} fingerprints are computed for toxic molecule SMILES, and Tanimoto similarity~\cite{tanimoto1958elementary} is used to measure structural proximity; subsequently, Butina clustering~\cite{butina1999unsupervised} is performed with a distance threshold of $0.4$ (corresponding to an intra-cluster similarity $\geq 0.6$, which is a commonly adopted choice in the community and consistent with recommended practices in the RDKit documentation)~\cite{evangelista2025application, landrum2013rdkit}. Given a target sample size $n$ and the number of clusters $k$, when $k \geq n$, one centroid molecule is selected from each of the largest $n$ clusters; when $k < n$, all centroid molecules are selected first, and the remaining samples are filled by boundary molecules within clusters to enhance structural coverage. Fig.~\ref{fig:dataset} shows the number of toxic molecules and corresponding clusters for each task. Meanwhile, Uniform Manifold Approximation and Projection (UMAP)~\cite{mcinnes2018umap} is used to project high-dimensional molecular representations into a 2D space for evaluating the structural coverage of the selected samples.

\begin{figure*}[t]
    \centering
    \includegraphics[width=\textwidth]{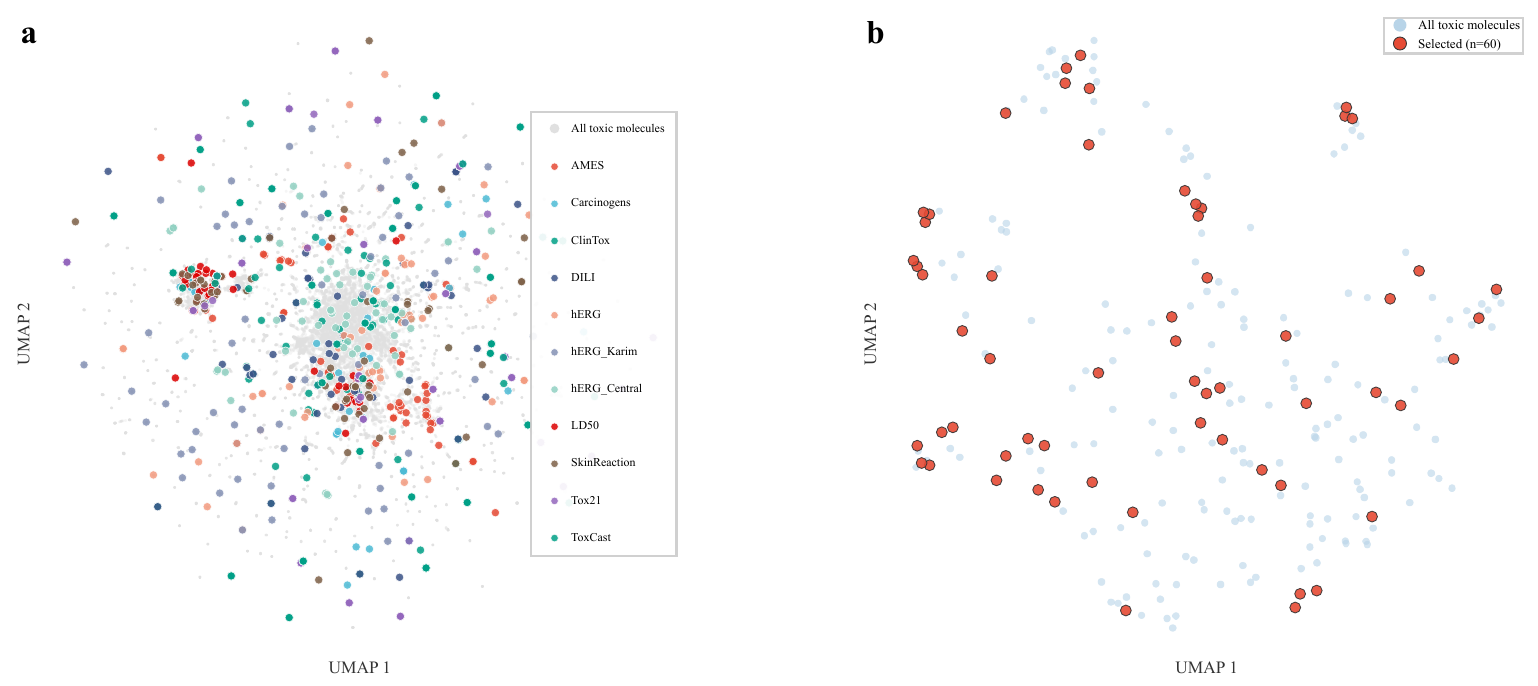}
    \caption{(a) Global chemical space coverage. UMAP embeddings are computed from ECFP4 fingerprints, where gray points denote all toxic molecules in the source datasets and colored points denote the 660 representative molecules selected via cluster-based sampling (colored by task). (b) A task-level example of cluster-based sampling (\texttt{DILI}). Light-blue points denote all 236 toxic molecules in the task, and red points denote the selected 60 representative molecules; the dispersed distribution reflects the diversity-oriented sampling strategy induced by Butina clustering.}
    \label{fig:umap_sampling}
\end{figure*}

To verify the representativeness and completeness of the adopted sampling strategy in structural space, we visualize the sampled molecules in chemical space. As shown in Fig.~\ref{fig:umap_sampling}~(a), the selected samples exhibit a relatively uniform distribution over the global chemical space, indicating that the strategy effectively covers the structural diversity of the original toxic molecules. Furthermore, Fig.~\ref{fig:umap_sampling}~(b) presents an example of fingerprint-based clustering and sampling for the DILI task, where the dispersed distribution of the sampled molecules in chemical space further confirms the ability of the proposed strategy to preserve structural diversity in the single-task setting.

In total, the ToxiMol benchmark contains $660$ toxic molecules (in SMILES format) and corresponding 2D molecular images, which are rendered using RDKit to construct multimodal inputs. It is worth noting that, as a benchmark designed for molecular toxicity repair, ToxiMol requires task-specific annotation pipelines for each primary task and its subtasks, and further necessitates structure-level validity checks and multi-endpoint toxicity evaluation for each generated repair candidate, constituting a typical high-cost, instance-level benchmark. Under this setting, the scale of $660$ molecules is comparable to existing benchmarks for complex molecular tasks targeting LLMs/MLLMs. For example, the MolPuzzle benchmark~\cite{guo2024can} contains $234$ challenge instances focused on structural reasoning; the USNCO-V benchmark~\cite{cui2025evaluating} collects $473$ multimodal chemistry problems; the SUPERChem benchmark~\cite{zhao2025superchem} constructs $500$ expert-curated chemical reasoning samples; the ChemVLM work further proposes MMChemBench~\cite{li2025chemvlm}, which includes $700$ multimodal questions for evaluating MLLMs in the chemistry domain.

\subsection{Prompt Annotation Pipeline}\label{prompt_pipeline}

Traditional MLLM benchmarks typically rely on generic prompt templates to guide model behavior, suitable for tasks with well-defined input-output structures and minimal mechanistic variation~\cite{hu2023promptcap, jin2022good, lan2023improving}. However, in the molecular toxicity repair task, models must address nearly 30 heterogeneous toxicity mechanisms and often struggle to capture the complex semantic relationships between mechanisms and repair objectives. A single template is insufficient to cover this intricate task space effectively. To address this, we propose a mechanism-aware prompt annotation pipeline that enables explicit alignment among toxicity mechanisms, molecular structures, and repair objectives.

Specifically, we design a prompt annotator $\mathcal{A}_{\text{prompt}}$ that dynamically transforms a general template into molecule-specific repair instructions. Given the task name $\mathcal{T}$ and a set of toxicity-positive SMILES strings $\mathcal{S}^{\text{raw}} = \{ \text{SMILES}_i^{\text{raw}} \}_{i=1}^{n}$, the annotator outputs a corresponding set of prompts $\mathcal{P} = \{ \mathcal{P}_i \}_{i=1}^{n}$, formalized as $\mathcal{A}_{\text{prompt}} : (\mathcal{T}, \mathcal{S}^{\text{raw}}) \rightarrow \mathcal{P}$. The annotation process consists of three stages: (1) loading a base template $\mathcal{P}_{\text{base}}$ that defines the MLLM role, repair objectives, and structural constraints; (2) injecting task-level annotations $\mathcal{P}_{\text{task}}$ and optional subtask-specific instructions $\mathcal{P}_{\text{subtask}}$; and (3) assembling, for each molecule $i$, a multimodal prompt by integrating its SMILES string $\text{SMILES}_i^{\text{raw}}$ and structure image $\text{IMG}_i$, resulting in $\mathcal{P}_i = \texttt{Assemble}(\mathcal{P}_{\text{base}}, \mathcal{P}_{\text{task}}, \mathcal{P}_{\text{subtask}}, \text{SMILES}_i^{\text{raw}}, \text{IMG}_i)$. The generated prompt is then passed to the model via $ \mathcal{M}(\mathcal{P}_i)$ to produce repair candidates. A complete example and scheduling algorithm are detailed in Appendix~\ref{annotation}.

\subsection{A Multi-Criteria Evaluation Chain}\label{ToxiEval}

\begin{figure*}[t]
  \centering
  \includegraphics[width=\textwidth]{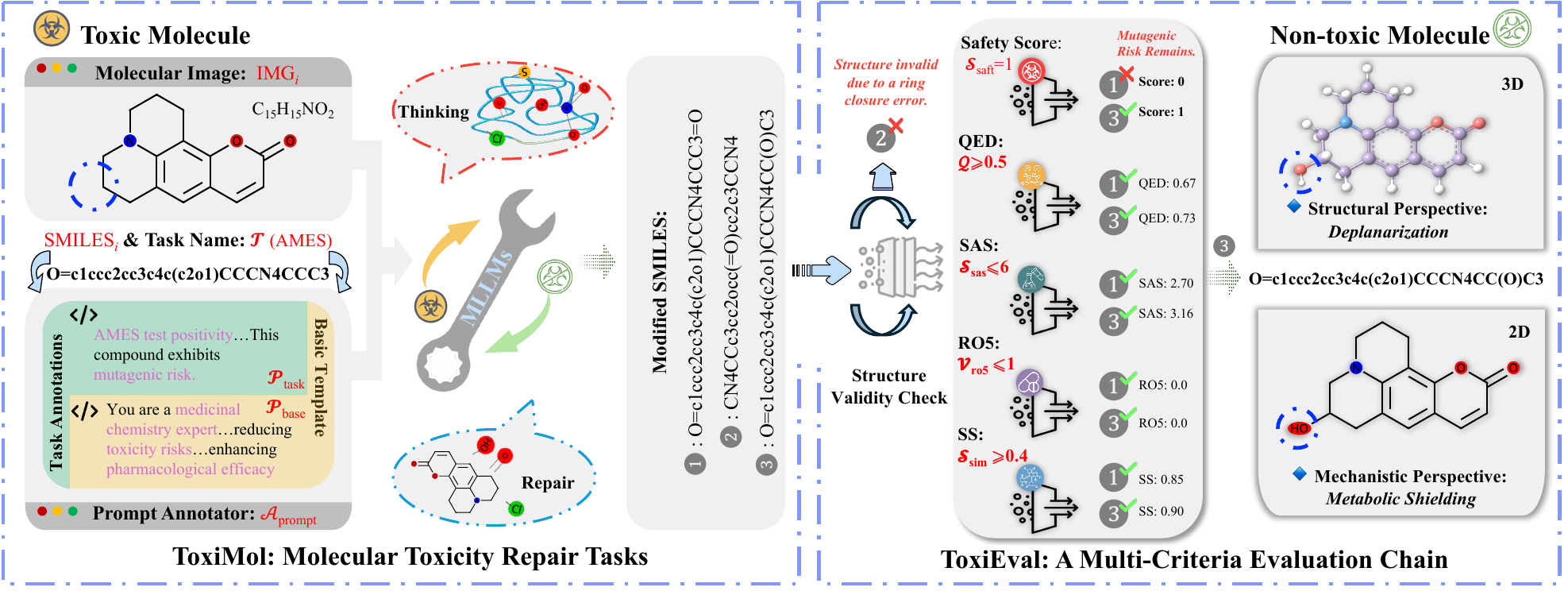}
  \caption{A repair example on the AMES task (binary classification, $\mathcal{S}_{\text{safe}} = 1$) using the ToxiMol benchmark. The input includes the SMILES of Coumarin 6H~\cite{pubchem94022}, its 2D molecular image, and a task-adaptive prompt annotation pipeline. After MLLM-mediated toxicity repair, three candidate molecules are generated. Due to structural invalidity, the ToxiEval evaluation chain first filters out Candidate 2. Among the remaining molecules, Candidate 1 is excluded because it fails the safety threshold in the multi-criteria assessment. Candidate 3 passes all stages of the ToxiEval chain and is identified as a successful repair.}
  \label{fig:ToxiMol_show}
\end{figure*}

The molecular toxicity repair task requires MLLMs to identify toxic substructures and propose structural modification strategies that not only reduce toxicity risks but also preserve desirable properties such as drug-likeness and synthetic feasibility. Unlike traditional classification tasks, toxicity repair does not have a single ground-truth answer in chemical space, and its success is determined by multiple synergistic objectives, making it unsuitable to evaluate using accuracy or other single-point metrics~\cite{li2021adversarial, manas2024improving}. To address this, we introduce \textit{ToxiEval}—a multi-criteria evaluation chain tailored for toxicity repair scenarios. ToxiEval encompasses key evaluation dimensions, including structural validity, safety scoring, and drug-likeness, among others, aiming to systematically and assess the real-world performance of general-purpose MLLMs on complex repair tasks. The overall execution flow of ToxiEval is illustrated in Fig.~\ref{fig:ToxiMol_show}, with the corresponding procedure given in Algorithm~\ref{alg:toxieval}. Additional case examples are provided in Appendix~\ref{successful_cases}.

\paragraph{Structure Validity Check.} Since the SMILES strings generated by general-purpose MLLMs do not always conform to syntactic standards, ToxiEval begins by invoking RDKit~\cite{rdkit} to parse each candidate molecule. The molecule is deemed invalid and excluded from subsequent evaluation if the SMILES cannot be converted into a valid molecular graph. This step is positioned at the beginning of the evaluation pipeline to ensure the credibility of the results and to save computational resources.

\paragraph{Multi-Dimensional Evaluation Criteria and Decision Protocol.} ToxiEval defines five core evaluation dimensions: Safety Score~$\mathcal{S}_{\text{safe}}$, Quantitative Estimate of Drug-likeness (QED)~$\mathcal{Q}$, Lipinski’s Rule of Five (RO5)~$\mathcal{V}_{\text{ro5}}$, Synthetic Accessibility Score (SAS)~$\mathcal{S}_{\text{sas}}$, and Structural Similarity (SS)~$\mathcal{S}_{\text{sim}}$. See Appendix~\ref{appendix_metrics} for details on these evaluation metrics. We adopt a strict conjunctive decision protocol, where a generated molecule is deemed \textit{successfully repaired} only if all criteria are satisfied simultaneously. The definitions and thresholds are as follows:

\begin{itemize}[leftmargin=15pt]
    \item \textbf{Safety Score:} We use the TxGemma-Predict~\cite{wang2025txgemma} model to estimate toxicity endpoints. For binary classification tasks, label ``A'' (non-toxic/non-inhibitory) corresponds to $\mathcal{S}_{\text{safe}} = 1$, while label ``B'' (toxic/inhibitory) corresponds to $\mathcal{S}_{\text{safe}} = 0$, and the threshold is $\mathcal{S}_{\text{safe}} = 1$. For the LD50 task, the model outputs a continuous score in $[0, 1000]$, where higher values indicate lower toxicity. This score is normalized to $[0, 1]$, and the threshold is set to $\mathcal{S}_{\text{safe}} > 0.5$, consistent with the default decision boundary of TxGemma-Predict. It is worth noting that our adoption of TxGemma-Predict as the sole molecular toxicity predictor is based on a comprehensive consideration of multiple factors. The selection criteria and their rationale have been systematically and rigorously discussed in Appendix~\ref{txgemma_selection}.

    \item \textbf{QED:} This criterion evaluates drug-likeness. QED outputs a continuous score in $[0, 1]$, with higher values indicating a stronger resemblance to real drug compounds. Following Bickerton \emph{et al}.~\cite{bickerton2012quantifying}, the threshold is $\mathcal{Q} \geq 0.5$.

    \item \textbf{SAS:} SAS assesses the synthetic feasibility of molecules, with scores ranging from $[1, 10]$. Lower values indicate easier synthesis. Based on Ertl and Schuffenhauer's findings, we set the threshold as $\mathcal{S}_{\text{sas}} \leq 6$~\cite{ertl2009estimation}.

    \item \textbf{RO5:} RO5 measures compliance with classic oral drug design rules based on properties such as molecular weight, LogP, and hydrogen bond donor/acceptor counts. As proposed by Lipinski \emph{et al}.~\cite{lipinski2012experimental}, the threshold is $\mathcal{V}_{\text{ro5}} \leq 1$.

    \item \textbf{SS:} Tanimoto similarity measures scaffold preservation between original and repaired molecules. Based on Rogers and Hahn, we set the threshold as $\mathcal{S}_{\text{sim}} \geq 0.4$~\cite{rogers2010extended}.
\end{itemize}

It should be noted that the thresholds for the above evaluation metrics are determined by consensus in the pharmacology community and authoritative references, and are supported by clear theoretical and practical foundations. Therefore, we consider it unnecessary to justify the rationality of these thresholds through additional sensitivity experiments.

\paragraph{Evaluation Algorithm.}\label{ToxiEval_algorithm}
Algorithm~\ref{alg:toxieval} presents the ToxiEval evaluation pipeline to determine whether a candidate molecule satisfies the multi-criteria toxicity repair constraints defined above; its notation is summarized in Appendix~\ref{ToxiEval_symbol}. Given an original molecule $\text{SMILES}_i^{\text{raw}}$ and a set of candidate repairs $\mathcal{S}_i^{\text{rep}}$ generated by an MLLM, the system first checks structural validity using RDKit. For valid candidates, it invokes the toxicity predictor $\mathscr{G}_{\text{tox}}$ to assign a safety score, either through label classification or normalized LD50 regression. Additional properties—including Drug-Likeness (QED), Synthetic Accessibility Score (SAS), Lipinski's Rule of Five (RO5), and Structural Similarity (SS)—are computed using cheminformatics tools. The evaluation adopts a strict conjunctive decision protocol: a candidate is deemed successful only if all criteria pass their respective thresholds. If any candidate in $\mathcal{S}_i^{\text{rep}}$ passes, the molecule is labeled as successfully repaired. This pipeline ensures rigorous, fine-grained assessment of molecular repair quality under realistic multi-objective constraints.

\begin{algorithm}[!htbp]
\small
\caption{\textbf{ToxiEval: Multi-Criteria Evaluation Pipeline for Toxicity Repair}}
\label{alg:toxieval}
\begin{algorithmic}[1]
\INPUT Candidate set $\mathcal{S}_i^{\text{rep}}=\{\text{SMILES}_i^{\text{rep}(1)},\text{SMILES}_i^{\text{rep}(2)},\text{SMILES}_i^{\text{rep}(3)}\}$; original molecule $\text{SMILES}_i^{\text{raw}}$; task identifier $\mathcal{T}$; toxicity predictor $\mathscr{G}_{\text{tox}}$; threshold set $\mathcal{C}^{\ast}$
\OUTPUT Repair success label $\texttt{success}_i \in \{0,1\}$

\STATE $\texttt{success}_i \leftarrow 0$

\FORALL{$s \in \mathcal{S}_i^{\text{rep}}$}
    \STATE \COMMENT{Step 1: Structure validity check}
    \IF{$\texttt{RDKitParse}(s) = \texttt{False}$}
        \STATE \textbf{continue}
    \ENDIF

    \STATE \COMMENT{Step 2: Safety scoring via toxicity prediction}
    \IF{$\mathcal{T} = \texttt{LD50}$}
        \STATE $v_{\text{tox}} \leftarrow \mathscr{G}_{\text{tox}}(s)$
        \STATE $\mathcal{S}_{\text{safe}} \leftarrow \texttt{Normalize}(v_{\text{tox}}, [0,1000])$
        \STATE $\texttt{passed}_{\text{safe}} \leftarrow (\mathcal{S}_{\text{safe}} > 0.5)$
    \ELSE
        \STATE $y_{\text{tox}} \leftarrow \mathscr{G}_{\text{tox}}(s)$
        \STATE $\texttt{passed}_{\text{safe}} \leftarrow (y_{\text{tox}} = \texttt{A})$
    \ENDIF

    \STATE \COMMENT{Step 3: Compute molecular evaluation metrics}
    \STATE $\mathcal{Q} \leftarrow \texttt{ComputeQED}(s)$
    \STATE $\mathcal{S}_{\text{sas}} \leftarrow \texttt{ComputeSAS}(s)$
    \STATE $\mathcal{V}_{\text{ro5}} \leftarrow \texttt{CountRO5Violations}(s)$
    \STATE $\mathcal{S}_{\text{sim}} \leftarrow \texttt{TanimotoSimilarity}(s,\ \text{SMILES}_i^{\text{raw}})$

    \STATE \COMMENT{Step 4: Apply conjunctive success decision}
    \IF{$\texttt{passed}_{\text{safe}}
        \wedge \mathcal{Q} \ge \mathcal{C}_{\text{qed}}^{\ast}
        \wedge \mathcal{S}_{\text{sas}} \le \mathcal{C}_{\text{sas}}^{\ast}
        \wedge \mathcal{V}_{\text{ro5}} \le \mathcal{C}_{\text{ro5}}^{\ast}
        \wedge \mathcal{S}_{\text{sim}} \ge \mathcal{C}_{\text{sim}}^{\ast}$}
        \STATE $\texttt{success}_i \leftarrow 1$
        \STATE \textbf{break}
    \ENDIF
\ENDFOR

\RETURN $\texttt{success}_i$
\end{algorithmic}
\end{algorithm}

\section{Experiments}
\label{sec:experiments}

\subsection{Benchmarking Procedure}
\label{sec:Benchmarking}

We systematically evaluate the performance of mainstream closed- and open-source MLLMs across \texttt{11} primary toxicity repair tasks. Additionally, the \texttt{12} sub-tasks from \texttt{Tox21} and \texttt{10} from \texttt{ToxCast} are included in extended evaluations (see Appendix~\ref{Tox21_success} and \ref{Toxcast_success}). All experiments are conducted under a unified testing environment equipped with \texttt{8 NVIDIA H200 GPUs}. Each model operates under a single-round QA setting, generating a fixed \texttt{3} repair candidates per sample to balance evaluation cost and performance; the impact of varying the number of generated candidates is analyzed in Fig.~\ref{fig:candidate}. The \texttt{temperature} parameter is set to \texttt{0.7} uniformly across all MLLMs; default settings are used when this is not configurable. A unified evaluation criterion is applied: \textbf{a sample is successfully repaired if at least one candidate molecule passes the ToxiEval multi-dimensional evaluation chain.} The computation of $\mathcal{S}_{\text{safe}}$ is based on the \texttt{TxGemma-27B-Predict}~\cite{wang2025txgemma} model (for model size ablation, see Appendix~\ref{TxGemma}). In our experimental setup, we adopt greedy decoding and deterministic regularized matching strategies for this model to ensure that identical inputs yield consistent predictions. Under this setting, the entire \texttt{ToxiEval} evaluation chain can be regarded as a hard-threshold-based binary decision system; therefore, we do not perform multiple stochastic samplings or variance statistics on the outputs of ToxiEval. The implementation details of $\mathcal{Q}$, $\mathcal{V}_{\text{ro5}}$, $\mathcal{S}_{\text{sas}}$, and $\mathcal{S}_{\text{sim}}$ are provided in Appendix~\ref{appendix_licenses}. We report the per-task and overall success rates as the primary performance metrics.

This work evaluates 43 mainstream MLLMs, of which 39 are general-purpose models. To compare against science/chemistry domain-specific MLLMs, we additionally include the Intern-S1 (science-domain)~\cite{bai2025intern} and ChemVLM (chemistry-domain)~\cite{li2025chemvlm} families, for a total of 4 models. All four domain-specific models are open-sourced and protocol-compatible with ToxiMol. Other domain-specific MLLMs (e.g., GIT-Mol~\cite{liu2024git}, InstructMol~\cite{cao2025instructmol}, Mol-LLM~\cite{lee2025mol}, TinyChemVL~\cite{zhao2025tinychemvl}, ChemMLLM~\cite{tan2025chemmllm}, and ChemDFM-X~\cite{zhao2024chemdfm}) are excluded solely due to reproducibility or input-compatibility constraints (e.g., unavailable weights or incompatible inputs).

\subsection{Overall Performance}\label{performance}

\paragraph{Model-Level Insights.} As shown in Table~\ref{tab:MLLM_success}, the overall repair success rate of ToxiMol still poses a significant challenge for existing general-purpose MLLMs. Notably, reasoning-enhanced variants exhibit only limited gains relative to their standard counterparts within the same model family (e.g., GPT-o3~\cite{openai_o3_o4mini_2025} vs.\ GPT-5.2~\cite{openai_gpt52_2025}, Qwen3-VL Thinking vs.\ Instruct~\cite{bai2025qwen3vl}), suggesting that reasoning capability has not yet established a clear advantage in the current task setting. Open-source MLLMs exhibit a scale effect, but it is not monotonically increasing; as model scale increases, some models degrade in performance. Among domain-specific models, Intern-S1 (241B total / 28B active, MoE) and Intern-S1 Mini (8B, dense)~\cite{bai2025intern} achieve performance comparable to some top-tier closed-source general-purpose models, indicating that science-oriented domain pretraining exhibits stronger transferability for toxicity repair. ChemVLM-8B~\cite{li2025chemvlm} demonstrates a level of competitiveness comparable to that of models of similar scale, whereas ChemVLM-26B~\cite{li2025chemvlm} performs worse; this phenomenon may be related to differences in backbone model and instruction fine-tuning strategies (see Appendix~\ref{chemvlm}).

\begin{table*}[!t]
  \setlength{\tabcolsep}{3pt}
  \normalsize
  \centering
   \caption{Toxicity Repair Success Rates (\%). hERG$\text{C}$, hERG$\text{K}$, and Carc$_\text{L}$ refer to the \texttt{hERG\_Central}, \texttt{hERG\_Karim}, and \texttt{Carcinogens} tasks, respectively. \texttt{SkinRxn} denotes the \texttt{SkinReaction} task. }
    \label{tab:MLLM_success}
  \renewcommand{\arraystretch}{1.0}
  \resizebox{\linewidth}{!}{
\begin{tabular}{@{}lcccccccccccc@{}}
\toprule
\textbf{Models} & \textbf{LD50} & \textbf{hERG$_C$} & \textbf{Carc$_L$} & \textbf{Tox21} & \textbf{AMES} & \textbf{ToxCast} & \textbf{ClinTox} & \textbf{DILI} & \textbf{hERG$_K$} & \textbf{hERG} & \textbf{SkinRxn} & \textbf{Overall} \\
\midrule
\multicolumn{13}{c}{\small{\textbf{Domain-Specific MLLMs}}} \\
\midrule
ChemVLM 26B~\cite{li2025chemvlm} & 0.0 & 26.7 & 16.7 & 31.7 & 1.7 & 28.3 & 33.3 & 1.7 & 13.3 & 6.7 & 8.3 & 15.3 \\
ChemVLM 8B~\cite{li2025chemvlm} & 0.0 & 35.0 & 20.0 & 38.3 & 18.3 & 43.3 & 35.0 & 3.3 & 16.7 & 8.3 & 10.0 & 20.8 \\
Intern-S1~\cite{bai2025intern} & 0.0 & 68.3 & 43.3 & 53.3 & 26.7 & 48.3 & 50.0 & 3.3 & 40.0 & 20.0 & 25.0 & \cellcolor{red!15} 34.4 \\
Intern-S1 Mini~\cite{bai2025intern} & 0.0 & 70.0 & 35.0 & 45.0 & 23.3 & 55.0 & 45.0 & 3.3 & 23.3 & 16.7 & 23.3 & 30.9 \\
\midrule

\textbf{Average Success Rate} 
& 0.0 & 50.0 & 28.8 & 42.1 & 17.5 & 43.7 & 40.8 & 2.9 & 23.3 & 12.9 & 16.7 & 25.4 \\

\midrule
\multicolumn{13}{c}{\small{\textbf{Close Source MLLMs}}} \\
\midrule
GPT-5.2~\cite{openai_gpt52_2025} & 0.0 & 68.3 & 38.3 & 48.3 & 21.7 & 53.3 & 51.7 & 5.0 & 43.3 & 25.0 & 26.7 & 34.7 \\
GPT-5.1~\cite{openai_gpt51_2025} & 0.0 & 61.7 & 36.7 & 43.3 & 23.3 & 55.0 & 41.7 & 5.0 & 35.0 & 16.7 & 23.3 & 31.1 \\
GPT-4.1~\cite{openai2025gpt41} & 1.67 & 63.3 & 35.0 & 48.3 & 20.0 & 46.7 & 46.7 & 5.0 & 35.0 & 15.0 & 20.0 & 30.6 \\
GPT-o3~\cite{openai_o3_o4mini_2025} & 1.7 & 75.0 & 45.0 & 50.0 & 33.3 & 56.7 & 58.3 & 5.0 & 55.0 & 26.7 & 30.0 & 39.7 \\
GPT-4o~\cite{hurst2024gpt} & 0.0 & 60.0 & 36.7 & 50.0 & 18.3 & 53.3 & 48.3 & 3.3 & 30.0 & 8.3 & 20.0 & 29.9 \\
GPT-o4-mini~\cite{openai_o3_o4mini_2025} & 0.0 & 75.0 & 40.0 & 51.7 & 25.0 & 51.7 & 48.3 & 3.3 & 48.3 & 33.3 & 28.3 & 36.8 \\

Claude Opus 4.5~\cite{anthropic_claude_opus4_5_2025} & 0.0 & 93.3 & 46.7 & 51.7 & 35.0 & 56.7 & 53.3 & 8.3 & 63.3 & 35.0 & 33.3 & \cellcolor{red!15} 43.3 \\
Claude Sonnet 4.5~\cite{anthropic_claude_sonnet4_5_2025} & 0.0 & 83.3 & 41.7 & 55.0 & 26.7 & 56.7 & 53.3 & 5.0 & 45.0 & 35.0 & 40.0 & 40.2 \\
Claude 3.7 Sonnet~\cite{claude37sonnet2024} & 0.0 & 78.3 & 33.3 & 50.0 & 26.7 & 53.3 & 50.0 & 5.0 & 51.7 & 26.7 & 28.3 & 36.7 \\

Gemini 3 Pro~\cite{google_gemini3_pro_official_2025} & 0.0 & 68.3 & 41.7 & 46.7 & 30.0 & 50.0 & 43.3 & 5.0 & 36.7 & 16.7 & 31.7 & 33.6 \\
Gemini 2.5 Pro~\cite{gemini25proexp2025} & 0.0 & 80.0 & 38.3 & 50.0 & 33.3 & 55.0 & 55.0 & 6.7 & 58.3 & 40.0 & 33.3 & 40.9 \\
Gemini 3 Flash~\cite{google_gemini3_pro_official_2025} & 1.7 & 71.7 & 36.7 & 48.3 & 36.7 & 48.3 & 51.7 & 6.7 & 40.0 & 18.3 & 28.3 & 35.3 \\
Gemini 2.5 Flash~\cite{google_gemini2_5_flash_2025} & 1.7 & 88.3 & 35.0 & 55.0 & 30.0 & 61.7 & 55.0 & 8.3 & 53.3 & 33.3 & 33.3 & 41.4 \\

Grok 4~\cite{xai_grok4_2025} & 3.3 & 76.7 & 33.3 & 46.7 & 30.0 & 43.3 & 43.3 & 6.7 & 41.7 & 20.0 & 31.7 & 34.2 \\
Grok 2 Vision~\cite{xai_grok2_vision_2024} & 0.0 & 53.3 & 35.0 & 46.7 & 18.3 & 48.3 & 41.7 & 3.3 & 21.7 & 21.7 & 23.3 & 28.5 \\

GLM-4V Plus~\cite{GLM4VPlus2024} & 0.0 & 33.3 & 16.7 & 36.7 & 11.7 & 45.0 & 33.3 & 5.0 & 15.0 & 6.7 & 10.0 & 19.4 \\

Doubao Seed 1.6 Vision~\cite{bytedance_doubao_seed1_6_vision_2024} & 0.0 & 58.3 & 26.7 & 35.0 & 20.0 & 43.3 & 36.7 & 5.0 & 23.3 & 13.3 & 18.3 & 25.5 \\
Doubao 1.5 Thinking-Vision Pro~\cite{volcengine_doubao_1_5_thinking_vision_pro_2025} & 0.0 & 60.0 & 35.0 & 38.3 & 23.3 & 51.7 & 41.7 & 5.0 & 30.0 & 13.3 & 26.7 & 29.6 \\

Moonshot v1 128K Vision~\cite{team2025kimi} & 1.7 & 53.3 & 26.7 & 48.3 & 21.7 & 40.0 & 45.0 & 6.7 & 35.0 & 15.0 & 18.3 & 28.3 \\
Hunyuan Vision~\cite{tencent2024hunyuanvision} & 0.0 & 26.7 & 20.0 & 36.7 & 10.0 & 33.3 & 30.0 & 5.0 & 23.3 & 6.7 & 8.3 & 18.2 \\
\midrule

\textbf{Average Success Rate} 
& 0.6 & 66.4 & 34.9 & 46.8 & 24.8 & 50.2 & 46.4 & 5.4 & 39.2 & 21.3 & 25.7 & 32.9 \\

\midrule
\multicolumn{13}{c}{\small{\textbf{Open Source MLLMs}}} \\
\midrule
\rowcolor{gray!15} \multicolumn{13}{l}{$\blacktriangledown$ \emph{Scale $<14$B}} \\
DeepSeek-VL2 Tiny~\cite{wu2024deepseek} & 0.0 & 28.3 & 18.3 & 36.7 & 3.3 & 33.3 & 36.7 & 5.0 & 10.0 & 8.3 & 5.0 & 16.8 \\
Qwen2.5-VL 3B Instruct~\cite{bai2025qwen2} & 0.0 & 35.0 & 15.0 & 18.3 & 6.7 & 40.0 & 36.7 & 3.3 & 15.0 & 8.3 & 8.3 & 17.0 \\
LLaVA-OneVision 7B~\cite{li2024llava} & 0.0 & 28.3 & 15.0 & 41.7 & 3.3 & 33.3 & 40.0 & 3.3 & 13.3 & 5.0 & 8.3 & 17.4 \\
MiMo v2 Flash~\cite{xiaomimimo_mimo_v2_flash_2025} & 1.7 & 61.7 & 25.0 & 48.3 & 18.3 & 51.7 & 45.0 & 3.3 & 33.3 & 18.3 & 18.3 & 29.4 \\
InternVL 3.5 8B~\cite{wang2025internvl3} & 0.0 & 65.0 & 30.0 & 48.3 & 18.3 & 56.7 & 48.3 & 3.3 & 25.0 & 15.0 & 23.3 & \cellcolor{red!15} 30.3 \\
Qwen2.5-VL 7B Instruct~\cite{bai2025qwen2} & 0.0 & 1.7 & 0.0 & 1.7 & 1.7 & 8.3 & 0.0 & 3.3 & 5.0 & 0.0 & 3.3 & 2.3 \\

\rowcolor{gray!15} \multicolumn{13}{l}{$\blacktriangledown$ \emph{14B$\leq$Scale$<72$B}} \\

DeepSeek-VL2 Small~\cite{wu2024deepseek} & 0.0 & 26.7 & 16.7 & 40.0 & 1.7 & 41.7 & 36.7 & 5.0 & 15.0 & 3.3 & 5.0 & 17.4 \\
DeepSeek-VL2~\cite{wu2024deepseek} & 0.0 & 30.0 & 16.7 & 36.7 & 1.7 & 40.0 & 36.7 & 3.3 & 11.7 & 6.7 & 5.0 & 17.1 \\
Yi-Vision~\cite{ai2024yi} & 0.0 & 43.3 & 30.0 & 45.0 & 30.0 & 48.3 & 46.7 & 5.0 & 23.3 & 11.7 & 25.0 & 28.0 \\
InternVL 3.5 14B~\cite{wang2025internvl3} & 0.0 & 71.7 & 33.3 & 43.3 & 25.0 & 53.3 & 51.7 & 8.3 & 38.3 & 20.0 & 21.7 & \cellcolor{red!15} 33.3 \\
InternVL 3.5 38B~\cite{wang2025internvl3} & 0.0 & 60.0 & 35.0 & 46.7 & 21.7 & 56.7 & 41.7 & 3.3 & 36.7 & 10.0 & 28.3 & 30.9 \\
Qwen2.5-VL 32B Instruct~\cite{bai2025qwen2} & 0.0 & 1.7 & 0.0 & 5.0 & 0.0 & 8.3 & 1.7 & 0.0 & 1.7 & 0.0 & 1.7 & 1.8 \\
Qwen3-VL 30B-A3B Instruct~\cite{bai2025qwen3vl} & 0.0 & 33.3 & 28.3 & 38.3 & 15.0 & 33.3 & 40.0 & 1.7 & 13.3 & 8.3 & 10.0 & 20.2 \\

\rowcolor{gray!15} \multicolumn{13}{l}{$\blacktriangledown$ \emph{Scale $\geq72$B}} \\
LLaVA-OneVision 72B~\cite{li2024llava} & 0.0 & 38.3 & 26.7 & 40.0 & 11.7 & 35.0 & 35.0 & 3.3 & 21.7 & 10.0 & 18.3 & 21.8 \\
Qwen2.5-VL 72B Instruct~\cite{bai2025qwen2} & 1.7 & 36.7 & 30.0 & 41.7 & 21.7 & 48.3 & 36.7 & 3.3 & 26.7 & 8.3 & 21.7 & 25.2 \\
InternVL 3 78B~\cite{zhu2025internvl3} & 0.0 & 46.7 & 25.0 & 46.7 & 18.3 & 45.0 & 38.3 & 3.3 & 23.3 & 10.0 & 23.3 & 25.5 \\
InternVL 3.5 241B-A28B~\cite{wang2025internvl3} & 0.0 & 66.7 & 35.0 & 56.7 & 28.3 & 61.7 & 50.0 & 8.3 & 46.7 & 25.0 & 25.0 & \cellcolor{red!15} 36.7 \\
Qwen3-VL 235B-A22B Instruct~\cite{bai2025qwen3vl} & 1.7 & 56.7 & 30.0 & 48.3 & 23.3 & 45.0 & 45.0 & 3.3 & 23.3 & 15.0 & 25.0 & 28.8 \\
Qwen3-VL 235B-A22B Thinking~\cite{bai2025qwen3vl} & 3.3 & 51.7 & 43.3 & 40.0 & 25.0 & 51.7 & 45.0 & 3.3 & 38.3 & 10.0 & 26.7 & 30.8 \\

\midrule

\textbf{Average Success Rate} 
& 0.4 & 41.2 & 23.9 & 38.1 & 14.5 & 41.7 & 37.5 & 3.8 & 22.2 & 10.2 & 16.0 & 22.7 \\

\bottomrule
\end{tabular}}
\end{table*}

\paragraph{Task-Level Analysis.} MLLMs achieve relatively higher success rates on \texttt{hERG$_C$}, \texttt{Tox21}, \texttt{ToxCast}, and \texttt{ClinTox}, indicating that general-purpose MLLMs exhibit a certain degree of generalization capability in handling hERG blockade prediction and toxicity phenotype classification tasks. However, the success rates on the \texttt{LD50} and \texttt{DILI} tasks are consistently very low. As a representative structural regression task, \texttt{LD50} requires models to explicitly account for continuous dose--response relationships of molecular toxicity during generation, which poses higher challenges to structural understanding and mechanistic reasoning. The \texttt{DILI} task involves hepatic metabolism and systemic toxicity, whose underlying mechanisms are highly complex, making it difficult for existing models to construct accurate representations. In addition, it is worth noting that models perform significantly worse on the standard \texttt{hERG} task than on \texttt{hERG$_C$} and \texttt{hERG$_K$}, despite their close toxicological relationships. To this end, we further design targeted ablation studies for detailed analysis, as reported in Appendix~\ref{hERG_analysis}.

\subsection{Ablation Studies}

This section presents six ablation studies: a structure validity analysis, the effect of candidate molecule count on success rate, failure mode attribution, multi-dimensional metric combination analysis, multi-round experimental validation, and the impact of multimodal perception.

\paragraph{Structure Validity Analysis.} We evaluate the structural validity of repaired molecules generated by different MLLMs. The experiment includes $660$ test samples, with $3$ candidate molecules generated per sample, resulting in $1980$ generated molecules. We present the results for \texttt{Claude Opus 4.5}~\cite{anthropic_claude_opus4_5_2025}, \texttt{Claude Sonnet 4.5}~\cite{anthropic_claude_sonnet4_5_2025}, \texttt{InternVL 3.5 241B-A28B} and \texttt{InternVL 3.5 38B}~\cite{wang2025internvl3} in Fig.~\ref{fig:Validity_success_rates}; results for other models are provided in Appendix~\ref{Valid}.

\begin{figure}[t]
  \centering
  \begin{minipage}[t]{0.5\linewidth}
    \centering
    \includegraphics[width=\linewidth]{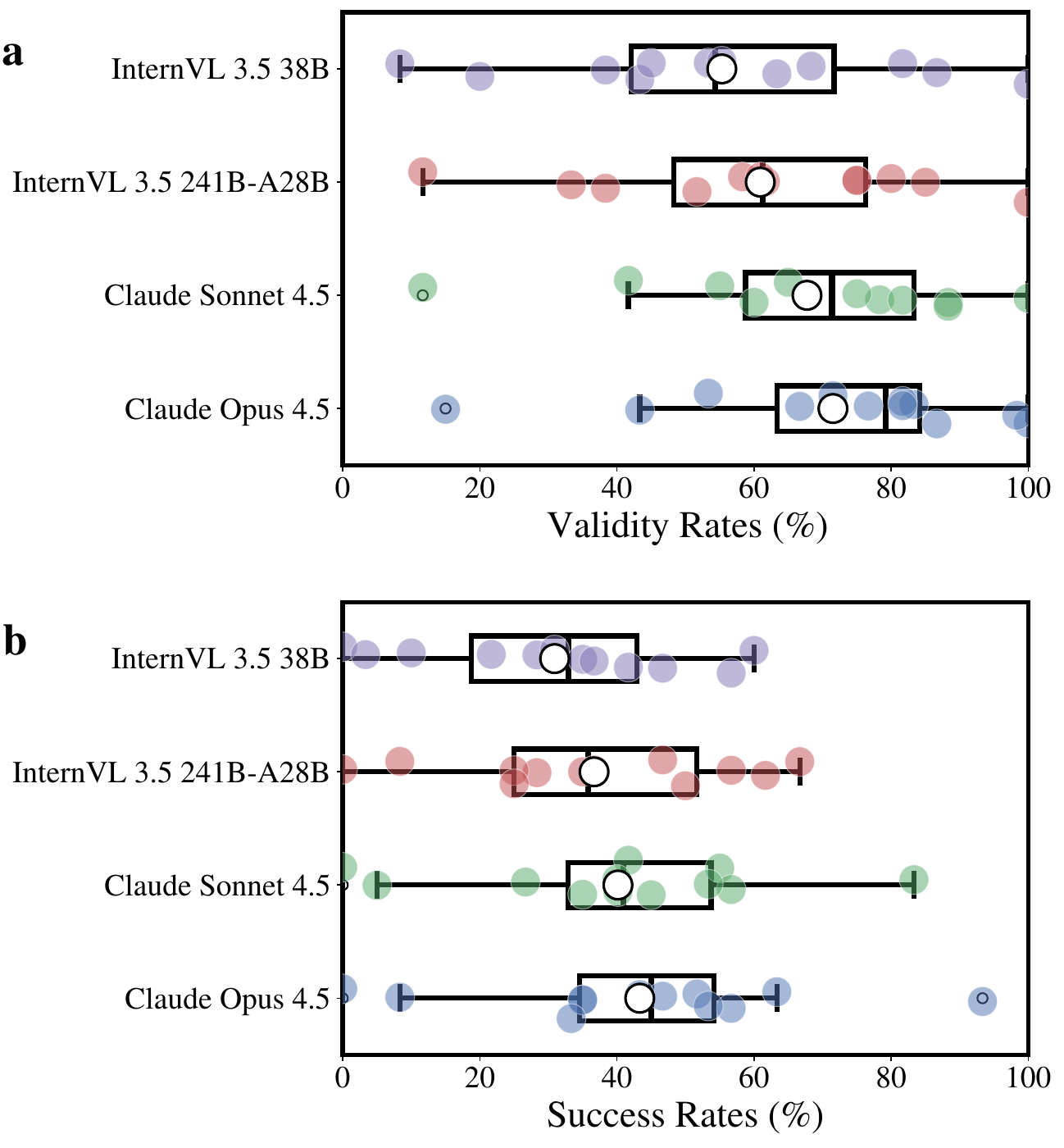}
    \caption{Distribution of validity and success rates~(\%) of representative MLLMs on the ToxiMol benchmark. Panel (a) shows the structural validity rates of each model, and panel (b) shows their repair success rates.}
    \label{fig:Validity_success_rates}
  \end{minipage}\hfill
  \begin{minipage}[t]{0.442\linewidth}
    \centering
    \includegraphics[width=\linewidth]{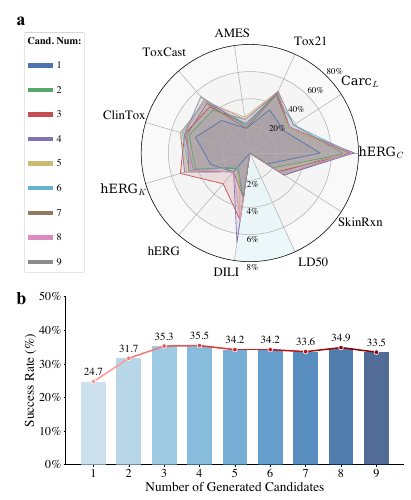}
    \caption{Effect of candidate count $k \in [1,9]$ on repair success rates. (a) Task-level rates; highlighted tasks use a 0--8\% scale (others: 0--80\%). (b) Overall rates versus $k$.}
    \label{fig:candidate}
  \end{minipage}
\end{figure}

From Fig.~\ref{fig:Validity_success_rates}, a clear distributional discrepancy between structural validity and repair success can be observed. In \textbf{panel (a)}, the validity distributions of all models are generally right-shifted, indicating that syntactically parsable molecular structures can be generated for most subtasks. In contrast, the success-rate distributions in \textbf{panel (b)} are overall substantially left-shifted and more dispersed, with some subtasks exhibiting near-zero values, suggesting that satisfying toxicity endpoint constraints is considerably more difficult than merely generating valid structures. Therefore, structural validity should be regarded as a necessary but insufficient condition for toxicity repair. Furthermore, the medians and dispersions of the box plots indicate differences in cross-subtask stability and overall performance among models: the \texttt{Claude} series exhibits higher and more concentrated distributions, whereas the \texttt{InternVL} series shows relatively lower performance with greater variability; within the same series, models with larger parameter scales tend to demonstrate certain advantages in both distribution level and stability. Overall, validity checking in \texttt{ToxiEval} is better suited as a preliminary filtering step to eliminate invalid structures, while the primary assessment of generation quality should rely on safety score and the satisfaction of multiple evaluation~criteria.

\paragraph{Effect of Candidate Molecule Count on Success Rate.} To systematically analyze the impact of the number of generated candidates $k$ on the success rate of toxicity repair, we select the \texttt{Claude-3.7 Sonnet}~\cite{claude37sonnet2024} model and vary $k \in [1,9]$ by dynamically specifying the number of molecules to be generated in the prompt. Each set of generated results is evaluated consistently using the ToxiEval framework. Fig.~\ref{fig:candidate} presents the task-level radar charts \textbf{panel (a)} and the corresponding overall success rates \textbf{panel (b)} under different values of $k$.

Experimental results show that increasing the number of candidates improves the success rate of toxicity repair, but with diminishing marginal returns. Specifically, the overall success rate rises from $24.70\%$ at $k{=}1$ to $\approx 35\%$ around $k{=}3$--$4$, and then plateaus with minor fluctuations for larger $k$. At the task level, different tasks exhibit varying sensitivity to $k$. For multi-label tasks such as \texttt{Tox21} and \texttt{ToxCast}, the success rate increases noticeably as $k$ grows in the low-$k$ regime and then approaches saturation, suggesting that additional candidate diversity mainly increases the chance of obtaining at least one feasible repair. In contrast, \texttt{LD50} remains at $0\%$ across all $k$, and \texttt{DILI} shows only limited improvement, indicating that merely increasing the number of candidates cannot overcome the intrinsic difficulty of certain toxicity endpoints under the current setting.

\paragraph{Failure Mode Attribution.}\label{failure_mode}
To diagnose why candidate molecules generated by MLLMs fail under \texttt{ToxiEval}, we quantify two representative failure modes among generated candidates. \textbf{Type-T} corresponds to RDKit-valid candidates that satisfy all non-toxicity constraints but fail the safety threshold, i.e., $\mathcal{S}_{\text{safe}} \neq 1$ for binary tasks or $\mathcal{S}_{\text{safe}} \leq 0.5$ for \texttt{LD50}. \textbf{Type-O} refers to RDKit-valid candidates that pass the safety threshold but still fail the overall repair criteria (i.e., at least one of $\mathcal{Q}$, $\mathcal{S}_{\text{sas}}$, $\mathcal{V}_{\text{ro5}}$, and $\mathcal{S}_{\text{sim}}$ is violated). Candidates that are RDKit-invalid or fail both the safety criterion and at least one non-toxicity constraint are not attributed to either type. Table~\ref{tab:typeT_typeO} reports the proportions (\%) of generated candidates attributed to Type-T and Type-O for \texttt{Claude-3.7 Sonnet}~\cite{claude37sonnet2024} across 11 tasks.

\begin{table}[t]
  \footnotesize
  \centering
  \caption{Failure-mode attribution for Claude-3.7 Sonnet on each toxicity-repair task. Percentages are computed over all generated candidate molecules for each task; Type-T/Type-O are assigned only to RDKit-valid candidates (others are not attributed to either type).}
  \label{tab:typeT_typeO}
  \renewcommand{\arraystretch}{1.25}
  \setlength{\tabcolsep}{5pt}

  \begin{tabularx}{\columnwidth}{@{}lcc>{\raggedright\arraybackslash}X@{}}
    \toprule
    \textbf{Task} &
    \textbf{Type-T (\%)} &
    \textbf{Type-O (\%)} &
    \textbf{Dominant Bottleneck} \\
    \midrule
    LD50              & \textbf{43.3} &  0.0 & Dose-toxicity regression bottleneck \\
    hERG$_{\text{C}}$ & \textbf{38.3} &  1.7 & Type-T dominant; Type-O marginal \\
    Carc$_{\text{L}}$ &  0.0 & \textbf{18.3} & Type-O dominant \\
    Tox21             &  3.3 & \textbf{23.3} & Type-O dominant \\
    AMES              & 11.7 & 13.3 & Mixed bottlenecks (Type-T $\approx$ Type-O) \\
    ToxCast           &  0.0 & \textbf{18.3} & Type-O dominant \\
    ClinTox           &  5.0 & 11.7 & Mild Type-O dominance \\
    DILI              & \textbf{41.7} &  0.0 & Strong toxicity bottleneck (Type-T) \\
    hERG$_{\text{K}}$ &  6.7 &  3.3 & Mild Type-T dominance \\
    hERG              & \textbf{38.3} &  1.7 & Type-T dominant \\
    SkinRxn           & 10.0 &  5.0 & Mild Type-T dominance \\
    \bottomrule
  \end{tabularx}

\end{table}

Overall, the table reveals a clear task-dependent stratification of bottlenecks. Toxicity bottlenecks dominate several endpoints, including \texttt{LD50}, \texttt{DILI}, and \texttt{hERG}, indicating that even when candidates satisfy drug-likeness constraints, meeting the safety criterion remains the primary hurdle. By contrast, drug-likeness bottlenecks are more pronounced in \texttt{Tox21} and \texttt{ToxCast}, suggesting that feasible toxicity mitigation may come at the cost of violating at least one drug-likeness or feasibility constraint. Several tasks exhibit mixed failure patterns, such as \texttt{AMES} and \texttt{hERG$_{\text{C}}$}, highlighting that failure modes can vary substantially across toxicity endpoints and evaluation constraints.

\paragraph{Multi-Dimensional Metric Combination Analysis.}\label{multi-dimensional_metric}
The \texttt{ToxiEval} framework integrates five evaluation dimensions: safety score $\mathcal{S}_{\text{safe}}$, drug-likeness $\mathcal{Q}$ (QED), synthetic accessibility $\mathcal{S}_{\text{sas}}$, rule-of-five violations $\mathcal{V}_{\text{ro5}}$, and scaffold similarity $\mathcal{S}_{\text{sim}}$. To quantify how different metric combinations shape the final success rate, we treat $\mathcal{S}_{\text{safe}}$ as a necessary condition and enumerate representative subsets of the remaining four constraints. We report results on four representative MLLMs, including \texttt{Claude 3.7 Sonnet}~\cite{claude37sonnet2024}, \texttt{Claude Opus 4.5}~\cite{anthropic_claude_opus4_5_2025}, \texttt{InternVL 3.5 8B}~\cite{wang2025internvl3}, and \texttt{InternVL 3.5 241B-A28B}~\cite{wang2025internvl3}. The overall success rates under each combination are summarized in Table~\ref{tab:criteria_ablation}.

\begin{table}[htbp]
  \centering
  \caption{Impact of Evaluation Metric Combinations on Success Rate.(\%)
  Each row corresponds to one metric-combination group: ``$k$-M$j$'' denotes a combination with $k$ enabled metrics (always including $\mathcal{S}_{\text{safe}}$) and the $j$-th variant among combinations of the same size; ``5-All'' enables all five metrics. \cmark/\xmark indicates whether a metric is enabled/disabled in the corresponding combination.}
  \label{tab:criteria_ablation}
  \renewcommand{\arraystretch}{1.08}
  \setlength{\tabcolsep}{3pt}

  \begin{adjustbox}{max width=\columnwidth}
  \begin{tabular}{c|ccccc|cccc}
    \toprule
    \multirow{2}{*}{\textbf{Group}} &
    \multicolumn{5}{c|}{\textbf{Combo}} &
    \multirow{2}{*}{\textbf{Claude 3.7 Sonnet}~\cite{claude37sonnet2024}} &
    \multirow{2}{*}{\textbf{Claude Opus 4.5}~\cite{anthropic_claude_opus4_5_2025}} &
    \multirow{2}{*}{\textbf{InternVL 3.5 8B}~\cite{wang2025internvl3}} &
    \multirow{2}{*}{\textbf{InternVL 3.5 241B-A28B}~\cite{wang2025internvl3}} \\
    \cmidrule(lr){2-6}
    & $\mathcal{S}_{\text{safe}}$ & $\mathcal{Q}$ & $\mathcal{S}_{\text{sas}}$ & $\mathcal{V}_{\text{ro5}}$ & $\mathcal{S}_{\text{sim}}$ & & & & \\
    \midrule
    1-M1 & \cmark & \xmark & \xmark & \xmark & \xmark & 58.3 & 62.4 & 45.9 & 51.8 \\
    \cmidrule(lr){1-10}
    2-M1 & \cmark & \cmark & \xmark & \xmark & \xmark & 41.8 & 49.4 & 33.3 & 39.4 \\
    2-M2 & \cmark & \xmark & \cmark & \xmark & \xmark & 57.6 & 63.0 & 45.5 & 50.9 \\
    2-M3 & \cmark & \xmark & \xmark & \cmark & \xmark & 55.9 & 60.3 & 43.9 & 50.2 \\
    2-M4 & \cmark & \xmark & \xmark & \xmark & \cmark & 52.4 & 56.1 & 42.9 & 48.2 \\
    \cmidrule(lr){1-10}
    3-M1 & \cmark & \cmark & \cmark & \xmark & \xmark & 41.8 & 49.4 & 33.2 & 39.4 \\
    3-M2 & \cmark & \cmark & \xmark & \cmark & \xmark & 41.8 & 49.4 & 33.3 & 39.4 \\
    3-M3 & \cmark & \cmark & \xmark & \xmark & \cmark & 36.7 & 43.3 & 30.5 & 36.7 \\
    \textbf{3-M4} & \cmark & \xmark & \cmark & \cmark & \xmark & \cellcolor{red!15} \textbf{55.9} & \cellcolor{red!15} \textbf{60.3} & \cellcolor{red!15} \textbf{43.9} & \cellcolor{red!15} \textbf{49.9} \\
    3-M5 & \cmark & \xmark & \cmark & \xmark & \cmark & 51.7 & 55.6 & 42.4 & 47.6 \\
    3-M6 & \cmark & \xmark & \xmark & \cmark & \cmark & 50.0 & 53.9 & 40.6 & 46.5 \\
    \cmidrule(lr){1-10}
    4-M1 & \cmark & \cmark & \cmark & \cmark & \xmark & 41.8 & 49.4 & 33.2 & 39.4 \\
    4-M2 & \cmark & \cmark & \cmark & \xmark & \cmark & 36.7 & 43.3 & 30.3 & 36.7 \\
    4-M3 & \cmark & \cmark & \xmark & \cmark & \cmark & 36.7 & 43.3 & 30.5 & 36.7 \\
    4-M4 & \cmark & \xmark & \cmark & \cmark & \cmark & 50.0 & 53.9 & 40.6 & 46.5 \\
    \cmidrule(lr){1-10}
    \textbf{5-All} & \cmark & \cmark & \cmark & \cmark & \cmark & \textbf{36.7} & \textbf{43.3} & \textbf{30.3} & \textbf{36.7} \\
    \bottomrule
  \end{tabular}
  \end{adjustbox}

  \vspace{0.2em}
\end{table}

Several consistent patterns emerge. First, using $\mathcal{S}_{\text{safe}}$ alone yields the highest success rate for all four models (Group 1-M1), indicating that toxicity reduction is the easiest objective to satisfy when no auxiliary constraints are imposed. Second, adding a single auxiliary constraint produces heterogeneous effects. In particular, introducing $\mathcal{Q}$ (Group 2-M1) consistently leads to the largest drop across models, whereas adding $\mathcal{S}_{\text{sas}}$ (Group 2-M2) or $\mathcal{V}_{\text{ro5}}$ (Group 2-M3) causes only mild degradation relative to 1-M1, suggesting that the generated candidates more frequently violate drug-likeness (as measured by QED) than basic feasibility (SAS) or RO5 constraints. Adding similarity $\mathcal{S}_{\text{sim}}$ alone (Group 2-M4) also reduces success, but the drop is typically smaller than that caused by $\mathcal{Q}$, implying that scaffold preservation is not the primary bottleneck compared with drug-likeness.

Third, when combining multiple auxiliary metrics, the success rate generally decreases monotonically as the constraint set becomes stricter, and the \textbf{full} five-metric configuration (Group 5-All) yields the lowest success across all models. Notably, the three-metric configuration $\{\mathcal{S}_{\text{safe}}, \mathcal{S}_{\text{sas}}, \mathcal{V}_{\text{ro5}}\}$ (Group 3-M4) achieves the best trade-off among the tested multi-metric subsets, matching the performance of $\mathcal{S}_{\text{safe}}+\mathcal{V}_{\text{ro5}}$ (Group 2-M3) and closely tracking $\mathcal{S}_{\text{safe}}+\mathcal{S}_{\text{sas}}$ (Group 2-M2) across models. This indicates that, under a fixed toxicity-reduction requirement, enforcing feasibility constraints via SAS and RO5 preserves most of the achievable repair rate while avoiding the sharp decline induced by adding $\mathcal{Q}$.

Finally, the four models exhibit similar qualitative trends but differ in their absolute sensitivity to constraint tightening. \texttt{Claude Opus 4.5}~\cite{anthropic_claude_opus4_5_2025} consistently attains the highest success under all combinations, while \texttt{InternVL 3.5 8B}~\cite{wang2025internvl3} remains the most sensitive to stricter constraint sets. Collectively, Table~\ref{tab:criteria_ablation} provides practical guidance on evaluation design: for rapid screening, the lightweight subset $\{\mathcal{S}_{\text{safe}}, \mathcal{S}_{\text{sas}}, \mathcal{V}_{\text{ro5}}\}$ offers a strong balance between success rate and feasibility, whereas comprehensive evaluation for downstream drug development should retain the complete five-dimensional constraint set.

\paragraph{Multi-Round Experimental Validation.}\label{mulit_round}
We acknowledge that MLLMs exhibit non-negligible stochasticity during generation. To partially mitigate this effect, \texttt{ToxiEval} adopts an ``any-candidate-succeeds'' rule: a sample is counted as successfully repaired if at least one generated candidate satisfies all evaluation constraints, which reduces the sensitivity of the final decision to random fluctuations. In addition, we perform a multi-round robustness check by selecting four representative MLLMs---two closed-source models (GPT-4.1~\cite{openai2025gpt41} and GPT-5.1~\cite{openai_gpt51_2025}) and two open-source models from the Qwen3-VL family (Qwen3-VL 30B-A3B Instruct~\cite{bai2025qwen3vl} and Qwen3-VL 235B-A22B Instruct~\cite{bai2025qwen3vl})---and repeating the full toxicity repair and evaluation pipeline five times under identical settings, reporting the mean and standard deviation.

\begin{table}[t]
  \centering
  \small
  \caption{Multi-round toxicity repair success rates (\%) across representative MLLMs (mean $\pm$ std over five runs). We report per-task and overall success rates for two closed-source models, GPT-4.1~\cite{openai2025gpt41} and GPT-5.1~\cite{openai_gpt51_2025}, and two open-source models from the Qwen3-VL family, Qwen3-VL 30B-A3B Instruct~\cite{bai2025qwen3vl} and Qwen3-VL 235B-A22B Instruct~\cite{bai2025qwen3vl}, evaluated under identical settings with five independent runs.}
  \label{tab:multi_round_results}
  \begin{tabular}{@{}lcccc@{}}
    \toprule
    & \multicolumn{2}{c}{\emph{Closed-Source MLLMs}} & \multicolumn{2}{c}{\emph{Open-Source MLLMs}} \\
    \cmidrule(lr){2-3}\cmidrule(l){4-5}
    \textbf{Task} & \textbf{GPT-4.1} & \textbf{GPT-5.1} & \textbf{Qwen3-VL 30B-A3B Instruct} & \textbf{Qwen3-VL 235B-A22B Instruct} \\
    \midrule
    LD50 & 0.3 $\pm$ 0.7 & 0.0 $\pm$ 0.0 & 0.0 $\pm$ 0.0 & 1.3 $\pm$ 0.7 \\
    hERG$_C$ & 61.0 $\pm$ 4.7 & 58.7 $\pm$ 3.7 & 30.7 $\pm$ 1.7 & 56.0 $\pm$ 1.3 \\
    Carc$_L$ & 34.0 $\pm$ 1.5 & 31.7 $\pm$ 3.2 & 29.3 $\pm$ 0.8 & 30.3 $\pm$ 0.7 \\
    Tox21 & 45.0 $\pm$ 1.8 & 44.3 $\pm$ 0.8 & 37.0 $\pm$ 1.9 & 48.7 $\pm$ 1.3 \\
    AMES & 20.0 $\pm$ 1.1 & 23.3 $\pm$ 1.1 & 17.7 $\pm$ 1.7 & 22.7 $\pm$ 1.4 \\
    ToxCast & 52.3 $\pm$ 4.2 & 55.0 $\pm$ 0.0 & 36.0 $\pm$ 2.7 & 46.7 $\pm$ 1.1 \\
    ClinTox & 47.7 $\pm$ 1.4 & 42.3 $\pm$ 2.3 & 41.0 $\pm$ 0.8 & 44.7 $\pm$ 0.7 \\
    DILI & 5.3 $\pm$ 0.7 & 3.3 $\pm$ 1.5 & 3.0 $\pm$ 1.3 & 3.3 $\pm$ 0.0 \\
    hERG$_K$ & 33.3 $\pm$ 4.6 & 32.3 $\pm$ 2.3 & 14.3 $\pm$ 1.3 & 24.3 $\pm$ 0.9 \\
    hERG & 14.3 $\pm$ 0.9 & 17.3 $\pm$ 0.8 & 6.7 $\pm$ 1.1 & 15.3 $\pm$ 1.3 \\
    SkinRxn & 21.7 $\pm$ 1.3 & 22.7 $\pm$ 3.1 & 12.0 $\pm$ 2.5 & 26.3 $\pm$ 2.0 \\
    \midrule
    Overall & 30.5 $\pm$ 0.7 & 30.1 $\pm$ 0.9 & 20.7 $\pm$ 0.6 & 29.1 $\pm$ 0.2 \\
    \bottomrule
  \end{tabular}
\end{table}

As shown in Table~\ref{tab:multi_round_results}, most task-level success rates exhibit limited variance across runs. The overall success rate is particularly stable, with standard deviations typically below 1 percentage point across all four models, indicating that the aggregated performance is robust to sampling noise even when certain individual tasks fluctuate mildly. Notably, the largest variance tends to appear on several relatively high-success tasks (e.g., \texttt{hERG$_C$} and \texttt{ToxCast}) for some models, whereas consistently hard endpoints such as \texttt{LD50} and \texttt{DILI} remain near-zero with minimal variation, suggesting that repeated sampling alone cannot overcome intrinsic task difficulty under the current setting. Given the substantial computational cost of repeating multi-round evaluations across all 43 MLLMs, we use single-round generation and evaluation results as the primary reporting scheme and provide this multi-round analysis as an auxiliary robustness validation.

\paragraph{Impact of Multimodal Perception.}\label{multimodal}
Although \texttt{ToxiMol} is designed as a multimodal benchmark, it remains unclear whether current MLLMs can effectively exploit 2D molecular images beyond the SMILES representation under our test-time prompting setup. To quantify the contribution of visual perception, we conduct a controlled ablation on \texttt{GPT-4.1}~\cite{openai2025gpt41} by comparing its standard multimodal setting (SMILES + 2D molecular image) against a text-only setting (SMILES only), while keeping all prompts, decoding configurations, and the \texttt{ToxiEval} evaluation procedure identical. Table~\ref{tab:gpt41_multimodal_text} reports the task-level and overall success rates (mean $\pm$ std over five runs) under the two settings.

\begin{table}[t]
  \centering
  \caption{Task-level success rates (\%) of multimodal and text-only models at test time. We compare \texttt{GPT-4.1}~\cite{openai2025gpt41} under the standard multimodal setting (SMILES + 2D molecular image) against a text-only variant (SMILES only). Reported values are mean $\pm$ std over five independent runs, evaluated under identical prompts, decoding settings, and the \texttt{ToxiEval} pipeline.}
  \label{tab:gpt41_multimodal_text}
  \footnotesize
  \setlength{\tabcolsep}{0pt}
  \begin{tabular*}{\linewidth}{@{\extracolsep{\fill}}l*{12}{c}@{}}
    \toprule
    \textbf{Setting} & \textbf{LD50} & \textbf{hERG$_C$} & \textbf{Carc$_L$} & \textbf{Tox21} & \textbf{AMES} & \textbf{ToxCast} & \textbf{ClinTox} & \textbf{DILI} & \textbf{hERG$_K$} & \textbf{hERG} & \textbf{SkinRxn} & \textbf{Overall} \\
    \midrule
    Multimodal & \meanstd{0.3}{0.7} & \meanstd{61.0}{4.7} & \meanstd{34.0}{1.5} & \meanstd{45.0}{1.8} & \meanstd{20.0}{1.1} & \meanstd{52.3}{4.2} & \meanstd{47.7}{1.4} & \meanstd{5.3}{0.7} & \meanstd{33.3}{4.6} & \meanstd{14.3}{0.9} & \meanstd{21.7}{1.3} & \meanstd{30.5}{0.7} \\
    \addlinespace[3pt]
    Text-only & \meanstd{0.3}{0.7} & \meanstd{56.3}{4.8} & \meanstd{34.3}{3.0} & \meanstd{45.0}{3.0} & \meanstd{16.3}{0.7} & \meanstd{51.0}{1.7} & \meanstd{44.3}{2.0} & \meanstd{4.7}{0.7} & \meanstd{31.0}{3.4} & \meanstd{12.3}{1.8} & \meanstd{22.0}{2.5} & \meanstd{28.9}{0.7} \\
    \bottomrule
  \end{tabular*}
\end{table}

Overall, multimodal perception yields a modest but consistent gain in the aggregated success rate, while the task-level effects vary across endpoints. For instance, improvements are more apparent on \texttt{hERG$_C$}, \texttt{AMES}, and \texttt{ClinTox}, whereas several tasks exhibit only marginal changes. These results suggest that simply providing molecular images does not guarantee substantial performance gains: current models may not reliably translate visual cues into structure-level edits that satisfy multi-criteria constraints. Importantly, this finding does not diminish the necessity of multimodal evaluation; rather, it highlights an open capability gap---\emph{effective utilization of molecular images for constrained structural repair remains challenging for contemporary MLLMs}. Consequently, \texttt{ToxiMol} provides a valuable testbed for future advances in multimodal alignment and representation learning that aim to better integrate visual perception with chemically grounded editing decisions.

\section{Conclusion}

This work systematically investigates the capability boundaries of general-purpose MLLMs on the \textit{molecular toxicity repair task} and proposes a new multimodal, multi-objective benchmark—\textbf{ToxiMol}. The benchmark covers many real-world toxicity mechanisms, provides a standardized dataset, and introduces a complementary multi-criteria evaluation framework, \textbf{ToxiEval}. Experimental results show that although existing MLLMs have not been explicitly optimized for this task, they already demonstrate preliminary potential for toxicity mitigation. 
\section{Limitations and Ethical Considerations} 

ToxiMol provides a unified and reproducible computational evaluation perspective. We do not equate the outputs of this surrogate predictor with definitive real-world toxicological correctness. Reliable toxicity conclusions still require wet-lab validation and multi-source evidence. Although ToxiMol covers diverse endpoints of toxicity repair, it remains extensible in both task scope and molecular modality. Future extensions may incorporate dose- and metabolism-aware constraints. The method may also move beyond small molecules to macromolecular therapeutics. Finally, while toxicity repair can support high-throughput early-stage screening, it also introduces non-trivial misuse risks, such as facilitating toxicity-evasive design of illicit compounds. This motivates careful ethical review, security auditing, and responsible release practices. For additional discussions, see Appendix~\ref{findings_appendix}.

\clearpage
\bibliographystyle{unsrtnat}
\bibliography{bibliography}

\hsappendix
\section{Rationale for Selecting the TxGemma-Predict Model}\label{txgemma_selection}

\subsection{Rationality and auditability of using an oracle as the evaluation coordinate system}

We adopt TxGemma-Predict~\cite{wang2025txgemma} as the unified computational oracle for the ToxiMol benchmark, balancing reproducibility and evaluation consistency. It is important to emphasize that molecular toxicity prediction itself inherently involves substantial uncertainty and task complexity in real-world drug discovery. In industrial and regulatory settings, high-confidence toxicity judgments still typically rely on wet-lab experiments and the integration of multi-source evidence, whereas computational predictions based on rule libraries, deep neural networks, or LLMs~\cite{greene2002computer, monem2025drug, wang2025txgemma} mainly provide approximate modeling of experimental outcomes under existing data and assumptions. Such predictors cannot guarantee complete unbiasedness or full coverage across all chemical spaces and all toxicity endpoints. Therefore, the goal of our evaluation is not to claim that the repaired molecules are absolutely non-toxic in the real world, but rather to examine whether the evaluated MLLMs can stably produce structure-level modification trends that \emph{reduce toxicity} under a unified, transparent, and reproducible computational standard. Under this positioning, TxGemma-Predict is used solely as a standardized reference oracle to measure changes in toxicity endpoints, and its outputs are not equated with ground-truth toxicological conclusions or wet-lab~labels.

At the implementation level, to ensure the auditability and reproducibility of the oracle, we fix the model version and inference configuration of TxGemma-Predict and adopt deterministic inference strategies, along with consistent post-processing and normalization rules. This ensures that identical inputs always produce exactly the same outputs under this oracle, allowing it to serve as a stable ``coordinate system'' for cross-model comparative evaluation. We acknowledge that, intuitively, adopting a computational oracle for evaluation may introduce unknown and unquantified systematic biases. However, from the historical development of AI benchmarks, the \emph{model-as-oracle} / \emph{model-as-judge} paradigm has numerous precedents in the standardized evaluation of highly complex tasks and has evolved into a mature, reproducible evaluation~methodology.

Specifically, in the LLM/MLLM benchmark literature, MT-Bench~\cite{zheng2023judging} employs an LLM as a judge to assess the preference and quality of dialogue model outputs; MM-Vet~\cite{yu2024mm} uses an LLM-based evaluator to assign unified scores to open-ended answers; and MMBench-Video~\cite{fang2024mmbench} similarly relies on LLMs to automatically evaluate free-form textual answers in video-based multimodal question answering. In the science benchmark domain, Practical Molecular Optimization (PMO)~\cite{gao2022sample} treats machine learning predictors fitted to experimental data as property oracles to evaluate molecular optimization methods under budget constraints; the molecular generation benchmarking platform MOSES~\cite{polykovskiy2020molecular} includes the Fréchet ChemNet Distance (FCD)~\cite{preuer2018frechet} metric, whose computation depends on the representation space of an external neural network, ChemNet~\cite{goh2018using}, to measure the discrepancy between generated and real distributions; and GuacaMol~\cite{brown2019guacamol} typically defines optimization objectives and performs standardized comparisons through reproducible scoring/oracle functions, which may include rule-based components, physicochemical property calculations, and learned predictors. Furthermore, at the level of general evaluation methodology, learned metrics constructed from pretrained model representations have long been standard tools for generation and understanding tasks. For example, FID~\cite{heusel2017gans} measures the distributional difference between generated and real images using features extracted by an Inception network and has been widely adopted as a generative model evaluation metric, alongside analogous metrics such as BERTScore~\cite{zhang2019bertscore}, BLEURT~\cite{sellam2020bleurt}, CLIPScore~\cite{hessel2021clipscore}, and FCD~\cite{preuer2018frechet}.

Taken together, these facts indicate that, for complex tasks lacking high-cost, fully covered human or experimental annotations, using a validated, version-controlled, and reproducible model as a standardized evaluation instrument is a practical approach widely adopted by the community over the long term. Based on this understanding, we believe that employing TxGemma-Predict as the unified toxicity oracle for ToxiMol and integrating it into the ToxiEval evaluation chain is methodologically justified.

\subsection{Selection of a high-quality, reproducible unified surrogate predictor}

We adopt TxGemma-Predict as the unified predictor for toxicity endpoints in the ToxiEval evaluation chain. The core rationale is that it achieves a relatively balanced trade-off among openness, reproducibility, cross-task consistency, and engineering usability, enabling it to serve as a standardized computational oracle in benchmark evaluation.

First, from the perspective of model capability and publicly available evidence, the original TxGemma paper reports that, on toxicity-related tasks, TxGemma demonstrates advantages over its predecessor, Tx-LLM~\cite{chaves2024tx}, and several representative molecular models (such as MolE~\cite{mendez2022mole} and LlaSMolMistral~\cite{yu2024llasmol}) on certain endpoints, while achieving comparable performance on others~\cite{wang2025txgemma}.

Second, in terms of coverage of the 11 toxicity endpoints involved in our benchmark, TxGemma is developed and systematically validated on the TDC/TxT~\cite{velez2024tdc, huang2021therapeutics, huang2022artificial} task collection. To the best of our knowledge, among publicly available, reproducible general-purpose toxicity prediction models, TxGemma and Tx-LLM are the only two models that can perform unified prediction across these 11 tasks using a single set of weights. Moreover, under the same evaluation settings, TxGemma exhibits superior overall performance, making it a more suitable surrogate oracle for benchmark evaluation.

Third, regarding the relationship to and comparability boundaries with public leaderboards, we note that the official TDC leaderboard provides verifiable reference results for several tasks (e.g., AMES: ZairaChem~\cite{turon2023first}; hERG: MapLight + GNN~\cite{notwell2023admet}; DILI: MiniMol~\cite{klaser2024minimol}; LD50: BaseBoosting~\cite{huang2022unified}). On these tasks, the performance of TxGemma-27B-Predict is on the same order of magnitude as the leading methods on the leaderboard and remains competitive. However, since different works may adopt different data versions and splitting strategies, these numerical results should not be interpreted as strict alignment with the current state of the art.

Fourth, from the perspective of engineering reproducibility and usage barriers, most top-performing methods on the TDC leaderboard are general modeling frameworks or general representation models rather than approaches specifically designed for a single task. They typically require retraining or adaptation on task-specific data before use, and their implementations depend on heterogeneous dependencies, feature engineering pipelines, and deployment environments. As a result, the overall reproduction cost is high and inconsistent. In contrast, TxGemma-Predict, as a publicly available, unified inference model, can directly produce predictions for multiple endpoints with a fixed model version and a unified inference interface, better satisfying the benchmark requirements for uniformity, transparency, and low-barrier reproducibility.

Fifth, existing ADMET tools (such as ProTox~\cite{banerjee2018protox}, ADMETlab~\cite{xiong2021admetlab}, pkCSM~\cite{pires2015pkcsm}, admetSAR~\cite{yang2019admetsar}, toxCSM~\cite{de2022toxcsm}, and predhERG~\cite{braga2015pred}) indeed provide rich toxicity proxies and endpoint prediction capabilities. However, these tools are often based on different data sources, experimental protocols, and threshold definitions. As a result, their endpoint sets and label semantics are difficult to strictly align with the 11 specific task labels defined in our benchmark, leading to conflicting outputs across different predictors. Stitching together multiple heterogeneous tools into a benchmark evaluation chain would significantly increase the cost of reproducibility and introduce additional inconsistencies. In contrast, adopting a single, unified surrogate predictor better satisfies the benchmark requirements for consistency and reproducibility.

Sixth, from the perspective of external usage evidence, TxGemma has been incorporated as a structure-level \emph{in silico} prediction model in several high-quality studies for comparison or as a baseline. For example, Bergen et al.~\cite{bergen2025large} include TxGemma-27B among the structure-based \emph{in silico} models for benchmark comparison and generalization testing in DILI-related work; Jones et al.~\cite{jones2025dataset} use TxGemma-2B as a baseline for TDC tasks to compare with their proposed methods. These practices indicate that TxGemma has a certain degree of community usability and reproducibility.

Finally, given the factors above, we believe selecting TxGemma-Predict as the unified surrogate predictor for the ToxiMol benchmark is reasonable in terms of methodological positioning, engineering reproducibility, and community usability. We reiterate that ToxiMol provides a unified and reproducible computational evaluation perspective; we do not equate the outputs of this computational oracle with the complete correctness of real-world toxicological mechanisms, and definitive toxicity conclusions still require wet-lab experiments and the integration of multi-source evidence.

\section{Toxicity Mechanisms Covered by ToxiMol}\label{toxicity_mechanisms}

\subsection{Main Tasks Toxicity Mechanisms}\label{main_tasks}

\begin{figure*}[t]
    \centering
    \includegraphics[width=1\textwidth]{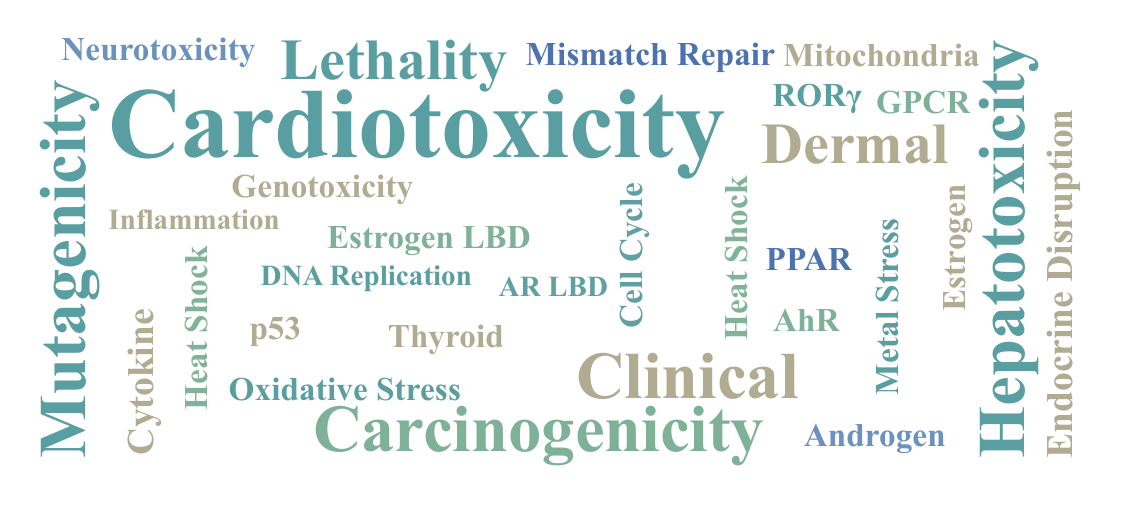}
    \caption{Word cloud visualization of the nearly 30 toxicity mechanisms covered by the main tasks and subtasks in~ToxiMol.}
    \label{fig:word_cloud}
\end{figure*}

The definitions of the main tasks and their corresponding toxicity mechanisms are summarized in Table~\ref{tab:tdc_tox_tasks_mechanisms}, where the mechanism descriptions for each task are compiled from the official \href{https://tdcommons.ai}{TDC documentation}. Furthermore, we visualize the nearly 30 toxicity mechanisms covered by the tasks, and the resulting word cloud of mechanism distributions is shown in Fig.~\ref{fig:word_cloud}.

\begin{table*}[!htbp]
\small
\centering
\caption{\textbf{Toxicity mechanism descriptions of the main tasks.}}
\label{tab:tdc_tox_tasks_mechanisms}
\begin{tabularx}{\textwidth}{@{\extracolsep{\fill}}l@{\hspace{60pt}}X}
\toprule
\textbf{Task} & \textbf{Toxicity Mechanism} \\
\midrule
AMES & Mutagenicity measured by Ames bacterial reverse-mutation assay (DNA damage/frameshift mutations). \\
hERG & Cardiotoxicity liability via blockade of the hERG potassium channel (QT prolongation/arrhythmia risk). \\
hERG\_Karim & hERG channel blockade liability (binary; thresholded around $<10\,\mu$M vs. $\ge 10\,\mu$M). \\
hERG\_Central & hERG channel inhibition profiling (percent inhibition/blocker classification from hERGCentral assays). \\
Tox21 & Target/pathway-specific in vitro toxicity across 12 assays (nuclear receptor signaling and stress-response pathways). \\
ToxCast & High-throughput toxicity screening across $>600$ in vitro assays spanning diverse biological targets/pathways. \\
Carcinogens & Carcinogenicity potential (cancer-promoting effects via genotoxicity and/or disruption of cellular metabolic processes). \\
SkinReaction & Immune-mediated skin sensitization from repeated exposure (allergic contact dermatitis; rLLNA-based). \\
DILI & Drug-induced liver injury (hepatotoxicity causing clinical liver damage and safety-related attrition). \\
LD50 & Acute systemic toxicity quantified by lethal dose (LD$_{50}$; higher dose implies stronger lethality). \\
ClinTox & Clinical toxicity risk: drugs failing clinical trials due to toxicity vs. drugs with successful trials. \\
\bottomrule
\end{tabularx}
\end{table*}

\subsection{Tox21 Subtask Definitions and Toxicity Mechanisms}\label{tox21_subtasks}
Table~\ref{tab:tox21_subtasks} overviews the 12 subtasks from the Tox21 dataset, listing each task’s identifier (T1–T12), subtask name, and the corresponding toxicity mechanism. These subtasks cover many biological pathways, including oxidative stress, DNA damage, endocrine receptor activation, and gene expression regulation. In our evaluation framework, each subtask is assessed independently to systematically evaluate the capacity of MLLMs to understand and address diverse mechanisms of molecular toxicity.

\begin{table*}[!htbp]
\small
\centering
\caption{\textbf{List of Tox21 Subtasks and Corresponding Toxicity Mechanisms.}}
\label{tab:tox21_subtasks}
\begin{tabularx}{\textwidth}{@{\extracolsep{\fill}}lll}
\toprule
\textbf{Task ID} & \textbf{Subtask Name} &  \textbf{Toxicity Mechanism} \\
\midrule
T1  & Tox21\_SR\_ARE         & Oxidative stress response via Antioxidant Response Element (ARE) \\
T2  & Tox21\_SR\_p53         & Genotoxic stress-induced activation of p53 tumor suppressor pathway \\
T3  & Tox21\_NR\_Aromatase   & Endocrine disruption via inhibition of aromatase enzyme \\
T4  & Tox21\_SR\_ATAD5       & DNA replication stress signaling through ATAD5 gene expression \\
T5  & Tox21\_NR\_ER\_LBD     & Estrogen receptor Ligand-binding Domain (ER-LBD) activation \\
T6  & Tox21\_SR\_HSE         & Heat shock response through Heat Shock Element (HSE) signaling \\
T7  & Tox21\_NR\_AR          & Androgen receptor (AR) activation and antagonism \\
T8  & Tox21\_NR\_PPAR\_$\gamma$ & Modulation of Peroxisome Proliferator-Activated Receptor-$\gamma$ (PPAR-$\gamma$) \\
T9  & Tox21\_NR\_ER          & Full-length Estrogen Receptor (ER) activation \\
T10 & Tox21\_SR\_MMP         & Mismatch repair signaling (MMP-related genotoxicity) \\
T11 & Tox21\_NR\_AhR         & Aryl Hydrocarbon Receptor (AhR) binding and activation \\
T12 & Tox21\_NR\_AR\_LBD     & Ligand-binding Domain modulation of Androgen Receptor (AR-LBD) \\
\bottomrule
\end{tabularx}
\end{table*}

\subsection{ToxCast Subtask Definitions and Toxicity Mechanisms}\label{toxcast_subtasks}
Table~\ref{tab:toxcast_subtasks} lists 10 representative subtasks selected from the ToxCast dataset, encompassing diverse toxicity pathways such as mitochondrial dysfunction, immunosuppression, developmental toxicity, metal response, GPCR activation, and neurotoxicity. These tasks extend the evaluation beyond Tox21 and assess the generalization ability of MLLMs in handling complex and varied molecular toxicity mechanisms. Each subtask is evaluated independently within the ToxiEval framework to ensure granularity and interpretability.

\begin{table*}[!htbp]
\scriptsize
\small
\caption{\textbf{List of ToxCast Subtasks and Corresponding Toxicity Mechanisms.}}
\label{tab:toxcast_subtasks}
\begin{tabularx}{\textwidth}{@{}llX@{}}
\toprule
\textbf{Task ID} & \textbf{Subtask Name} & \textbf{Toxicity Mechanism} \\
\midrule
T1  & ToxCast\_APR\_HepG2\_MitoMass\_24h\_dn & Mitochondrial toxicity in HepG2 cells (24h exposure) \\
T2  & ToxCast\_BSK\_3C\_ICAM1\_down          & Suppression of ICAM1 expression in immune cell model (inflammation) \\
T3  & ToxCast\_Tanguay\_ZF\_120hpf\_TR\_up   & Zebrafish developmental response at 120 hpf (transcriptional activation) \\
T4  & ToxCast\_ATG\_M\_32\_CIS\_dn           & Downregulation of CIS gene (cell proliferation inhibition) \\
T5  & ToxCast\_ATG\_ROR$\gamma$\_TRANS\_up  & Activation of Retinoic Acid Receptor-related Orphan Receptor-$\gamma$ (ROR-$\gamma$) \\
T6  & ToxCast\_TOX21\_p53\_BLA\_p3\_ch2      & Genotoxic response via p53 reporter activation \\
T7  & ToxCast\_BSK\_3C\_MCP1\_down           & Immunosuppression via MCP1 downregulation in BioMAP assay \\
T8  & ToxCast\_ATG\_MRE\_CIS\_up             & Activation of Metal-Responsive Element (MRE) pathway \\
T9  & ToxCast\_NVS\_GPCR\_gLTB4              & GPCR pathway activation via Leukotriene B4 receptor binding \\
T10 & ToxCast\_NVS\_ENZ\_hAChE               & Inhibition of human acetylcholinesterase (neurotoxicity marker) \\
\bottomrule
\end{tabularx}
\end{table*}
\section{Supplementary Ablations and Analyses}

\subsection{Analysis of hERG-Related Tasks}\label{hERG_analysis}

In the overall benchmarking results (Section~\ref{performance}), we observe a persistent gap in success rates across the three hERG-related tasks (\texttt{hERG}, \texttt{hERG$_{\text{C}}$}, and \texttt{hERG$_{\text{K}}$}). To examine whether this gap is driven by prompt formulation rather than the toxicity endpoint itself, we conduct a controlled prompt ablation by unifying the prompt template for all three tasks. Concretely, we replace the original task-specific prompts with a single unified template (derived from the \texttt{hERG$_{\text{C}}$} prompt format) while keeping the test set, decoding settings, and the \texttt{ToxiEval} pipeline (including \texttt{TxGemma-Predict}~\cite{wang2025txgemma}) unchanged. \texttt{Claude-3.7 Sonnet}~\cite{claude37sonnet2024} is used as the representative generator. Table~\ref{tab:hERG_analysis} reports the success rates under the original versus unified prompt settings.

\begin{table}[htbp]
  \small
  \centering
  \caption{Comparison of success rates (\%) on hERG-related tasks using the original versus unified prompt templates. Results are obtained with \texttt{Claude-3.7 Sonnet}~\cite{claude37sonnet2024} as the repair generator, evaluated under the \texttt{ToxiEval} pipeline with \texttt{TxGemma-Predict}~\cite{wang2025txgemma}. The ``Unified'' setting applies a single shared prompt template (derived from the \texttt{hERG$_{\text{C}}$} format) to all three tasks while keeping the test set and decoding configuration unchanged.}
  \label{tab:hERG_analysis}
  \renewcommand{\arraystretch}{1.15}
  \setlength{\tabcolsep}{5pt}
  \begin{tabularx}{\linewidth}{@{}l*{3}{>{\centering\arraybackslash}X}@{}}
    \toprule
    \textbf{Prompt Template} & \textbf{hERG} & \textbf{hERG$_{\text{C}}$} & \textbf{hERG$_{\text{K}}$} \\
    \midrule
    Original & 26.7 & 78.3 & 51.7 \\
    Unified  & 26.7 & 81.7 & 53.3 \\
    \bottomrule
  \end{tabularx}
\end{table}

As shown in Table~\ref{tab:hERG_analysis}, prompt unification yields only marginal changes: the success rate on \texttt{hERG} remains unchanged, while \texttt{hERG$_{\text{C}}$} and \texttt{hERG$_{\text{K}}$} improve slightly. These results suggest that differences in prompt formatting are not the primary cause of the pronounced performance discrepancy between \texttt{hERG} and the other two variants. Instead, the gap is more likely attributable to differences in endpoint semantics and evaluation strictness under \texttt{TxGemma-Predict}. In particular, the \texttt{hERG} task is phrased in terms of ``block'' rather than ``inhibit'', which is commonly interpreted as a stricter toxicological criterion and may induce a harder safety decision boundary, thereby limiting the achievable repair success even when using a unified prompt template.

\subsection{TxGemma-Predict Model Selection}\label{TxGemma}

In \texttt{ToxiEval}, toxicity endpoint prediction is implemented using the \texttt{TxGemma-Predict} model family~\cite{wang2025txgemma}, which provides three parameter scales (2B, 9B, and 27B). To assess how the evaluator scale affects toxicity-repair outcomes, we keep all other settings fixed and re-evaluate the same set of repair candidates generated by \texttt{Claude-3.7 Sonnet}~\cite{claude37sonnet2024} under each TxGemma variant. Table~\ref{tab:txgemma_family} summarizes the resulting task-level and overall success rates.

\begin{table*}[htbp]
  \scriptsize
  \centering
  \caption{Task-level success rates (\%) under different \texttt{TxGemma-Predict} evaluator scales. We re-evaluate the same set of repair candidates using the \texttt{TxGemma-Predict} family~\cite{wang2025txgemma} (2B/9B/27B) as the toxicity oracle; all other settings are kept fixed.}
  \label{tab:txgemma_family}
  \renewcommand{\arraystretch}{1.2}
  \setlength{\tabcolsep}{3.5pt}
  \begin{adjustbox}{max width=\textwidth}
    \begin{tabular*}{\textwidth}{@{\extracolsep{\fill}}l|*{12}{c}@{}}
      \toprule
      \textbf{Model} &
      \textbf{LD50} & \textbf{hERG$_{\text{C}}$} & \textbf{Carc$_{\text{L}}$} &
      \textbf{Tox21} & \textbf{AMES} & \textbf{ToxCast} & \textbf{ClinTox} &
      \textbf{DILI} & \textbf{hERG$_{\text{K}}$} & \textbf{hERG} & \textbf{SkinRxn} &
      \textbf{Overall} \\
      \midrule
      TxGemma-Predict-2B  & 0.0 & 88.3 & 35.0 & 46.7 & 23.3 & 55.0 & 56.7 & 5.0 & 63.3 & 10.0 & 6.7 & 35.5 \\[2pt]
      TxGemma-Predict-9B  & 0.0 & 91.7 & 41.7 & 55.0 & 36.7 & 60.0 & 51.7 & 5.0 & 58.3 & 16.7 & 3.3 & 38.2 \\[2pt]
      TxGemma-Predict-27B & 0.0 & 93.3 & 46.7 & 51.7 & 35.0 & 56.7 & 53.3 & 8.3 & 63.3 & 35.0 & 33.3 & 43.3 \\[2pt]
      \bottomrule
    \end{tabular*}
  \end{adjustbox}
\end{table*}

As shown in Table~\ref{tab:txgemma_family}, increasing the evaluator capacity leads to consistent gains in overall success rate, with \texttt{TxGemma-Predict-27B} achieving the highest average performance (43.3\%) compared to the 9B (38.2\%) and 2B (35.5\%) variants. Notably, the improvements are task-dependent: \texttt{TxGemma-Predict-27B} yields substantial gains on \texttt{hERG} and \texttt{SkinRxn}, while differences on several other endpoints are moderate. In our implementation, the 27B variant is evaluated in full precision (\texttt{float32}) due to stability considerations at this scale, whereas the 2B and 9B variants use \texttt{float16} for efficiency. Unless otherwise stated, we therefore adopt \texttt{TxGemma-Predict-27B} as the default toxicity evaluator in \texttt{ToxiEval}, aligning the benchmark oracle with the strongest publicly available model in the TxGemma family.

\subsection{Analysis of ChemVLM-8B vs. ChemVLM-26B}\label{chemvlm}

As shown in Table~\ref{tab:MLLM_success}, ChemVLM-26B exhibits lower overall performance than ChemVLM-8B. We strictly followed the official deployment procedure and verified the model configurations: ChemVLM-8B uses InternLM2.5-7B-Chat~\cite{cai2024internlm2} as its language backbone, whereas the language component of ChemVLM-26B is based on the InternLM2-Chat-20B~\cite{cai2024internlm2} family, and the model card states that it is fine-tuned by combining ChemLLM-20B~\cite{zhang2024chemllm} and InternViT-6B~\cite{chen2024internvl}. Since InternLM2.5 has stronger instruction-following capability, ChemVLM-8B may be more robust under structured output constraints (e.g., outputting only SMILES). In addition, we further analyzed the outputs of ChemVLM-26B and found that it exhibits ultra-long repetitive segments and non-standard SMILES forms in several tasks (e.g., \texttt{CO[O-]/C=C/C=C/\ldots}), suggesting that it may be more prone to generation degeneration or more sensitive to decoding/output constraints in such structured generation settings; the underlying causes may be related to backbone differences, tokenizer/decoding strategies, and alignment focus. The 26B model may place more emphasis on general multimodal understanding (image captioning and chemistry QA), whereas the 8B model may be more optimized for structured output generation (SMILES).

\section{Supplementary Material for the Mechanism-Aware Prompt Annotation~Pipeline}\label{annotation}

\subsection{Prompt Template Example for Tox21 Tasks}\label{appendix_prompt_template}

\begin{figure*}[htbp]
  \centering
  \includegraphics[width=0.9\linewidth]{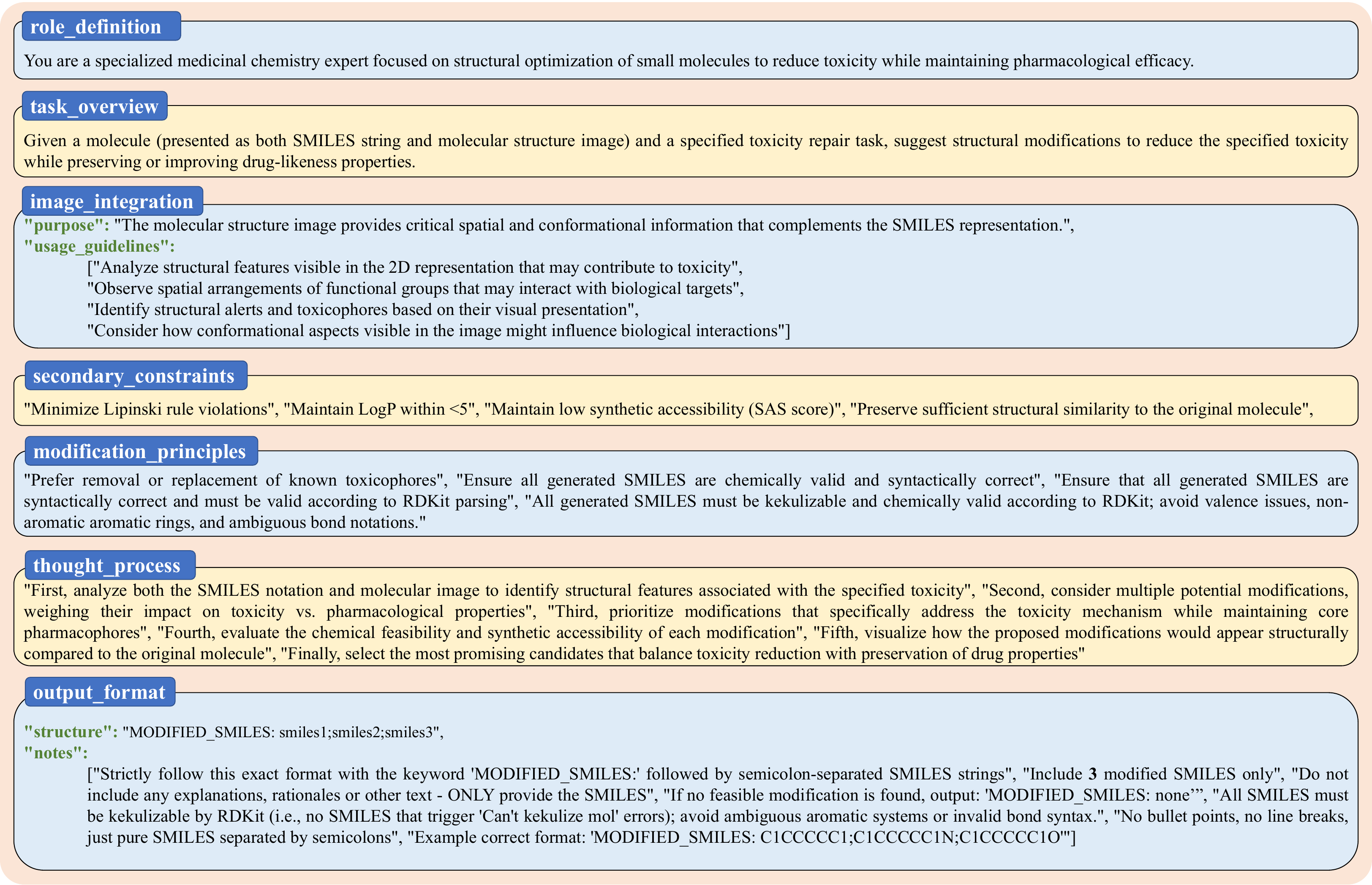}
  \caption{\textbf{Overall prompt design for detoxification tasks using MLLMs.}}
  \label{fig:prompt_overall}
\end{figure*}

\begin{figure*}[htbp]
  \centering
  \includegraphics[width=0.9\linewidth]{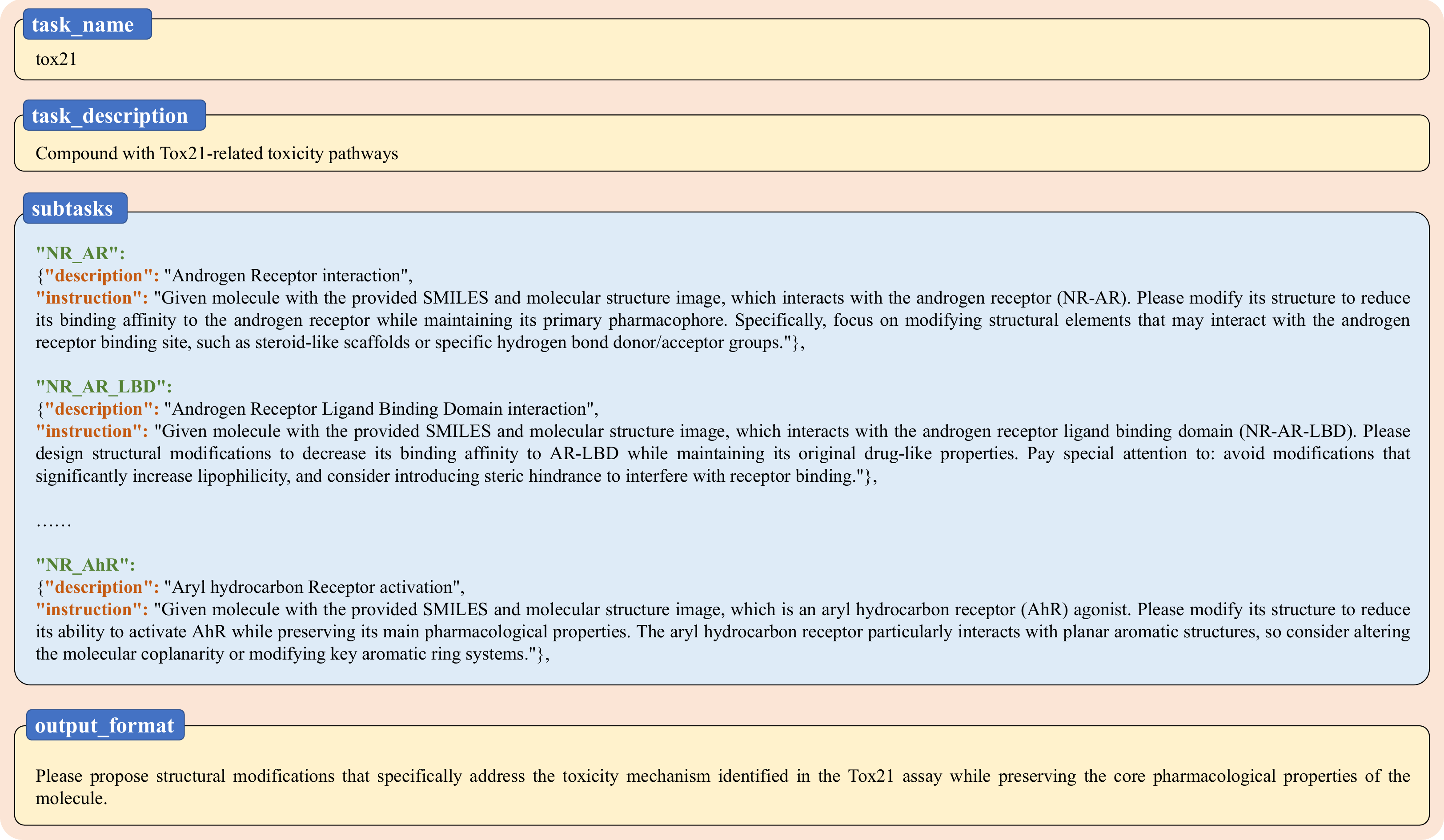}
  \caption{\textbf{Task prompt design for detoxification tasks of Tox21 using MLLMs.}}
  \label{fig:prompt_task}
\end{figure*}

In this appendix, we provide an illustrative template to complement the mechanism-aware prompt annotation pipeline introduced in Section~\ref{prompt_pipeline}. The included configuration consists of the base prompt template, which defines the MLLM role, repair objectives, and formatting constraints shown in Fig. \ref{fig:prompt_overall}, and task-level annotations specific to the Tox21 dataset shown in Fig. \ref{fig:prompt_task}. Furthermore, we include 12 subtask-specific prompt fragments corresponding to the individual endpoints within the Tox21 benchmark (T1–T12), as shown in Table~\ref{tab:tox21_subtasks}, each reflecting the unique toxicity mechanism it represents. This example serves as a reference for understanding how prompts are hierarchically composed and adapted to different toxicity~contexts.

\subsection{Symbolic Notation for Mechanism-Aware Prompt Annotation Pipeline}\label{prompt_symbolic}

Table~\ref{tab:prompt_symbol_definitions} summarizes the symbolic notations used throughout the mechanism-aware prompt annotation pipeline described in Subsection~\ref{prompt_pipeline}. These symbols represent the key components of the toxicity repair process, including task identifiers, molecule inputs, 2D structural images, hierarchical prompt components, and MLLM-generated outputs.

\begin{table*}[!htbp]
\small
\centering
\caption{\textbf{Symbol Definitions for Mechanism-Aware Prompt Annotation Pipeline.}}
\label{tab:prompt_symbol_definitions}
\begin{tabularx}{\textwidth}{@{\extracolsep{\fill}}ll}
\toprule
\textbf{Symbol} & \textbf{Definition} \\
\midrule
$\mathcal{T}$ & Toxicity repair task identifier (e.g., AMES, DILI, hERG). \\
$\mathcal{S}^{\text{raw}} = \{ \text{SMILES}_i^{\text{raw}} \}_{i=1}^{n}$ & Set of toxicity-positive SMILES strings requiring repair. \\
$\text{IMG}_i$ & 2D molecular structure image associated with $\text{SMILES}_i^{\text{raw}}$, rendered via RDKit. \\
$\mathcal{P}_{\text{base}}$ & Base prompt template encoding MLLM role, objectives, and formatting rules. \\
$\mathcal{P}_{\text{task}}$ & Task-level annotation describing toxicity mechanism and optimization focus. \\
$\mathcal{P}_{\text{subtask}}$ & Subtask-level annotation for fine-grained tasks (e.g., specific Tox21 endpoints). \\
$\mathcal{P}_i$ & Final prompt for molecule $i$, including template, annotations, SMILES, and image. \\
$\mathcal{M}$ & MLLMs queried for toxicity repair. \\
$\mathcal{S}_i^{\text{rep}} = \{ \text{SMILES}_i^{\text{rep}(k)} \}_{k=1}^3$ & Candidate repaired SMILES outputs generated by $\mathcal{M}$ for molecule $i$. \\
$\mathcal{S}^{\text{rep}}$ & Final set of all generated candidate molecules across the input batch. \\
\bottomrule
\end{tabularx}
\end{table*}

\subsection{Algorithm for Prompt Annotation Pipeline}\label{prompt_algorithm}

Algorithm~\ref{alg:mechanism_prompt} presents the mechanism-aware prompt annotation pipeline for generating molecule-specific repair instructions in the ToxiMol benchmark. Given a toxicity repair task $\mathcal{T}$, a set of toxicity-positive molecules $\mathcal{S}^{\text{raw}}$, and corresponding structure images $\{\text{IMG}_i\}$, the system first loads a shared base prompt $\mathcal{P}_{\text{base}}$ and task-level annotations $\mathcal{P}_{\text{task}}$. For tasks with subtask granularity, an additional subtask annotation $\mathcal{P}_{\text{subtask}}$ is injected per molecule based on its inferred category.

\begin{algorithm}[htbp]
\caption{\textbf{Mechanism-Aware Prompt Annotation Pipeline}}
\label{alg:mechanism_prompt}
\begin{algorithmic}[1]
\INPUT Toxicity repair task name $\mathcal{T}$; toxicity-positive molecule set $\mathcal{S}^{\text{raw}}=\{\text{SMILES}_i^{\text{raw}}\}_{i=1}^{n}$; corresponding images $\{\text{IMG}_i\}_{i=1}^{n}$; MLLM model $\mathcal{M}$
\OUTPUT Repaired molecule candidates $\mathcal{S}^{\text{rep}}$

\STATE Initialize empty output set: $\mathcal{S}^{\text{rep}} \leftarrow \emptyset$
\STATE Load base template: $\mathcal{P}_{\text{base}} \leftarrow \texttt{LoadBasePrompt}()$
\STATE Load task-level annotations: $\mathcal{P}_{\text{task}} \leftarrow \texttt{LoadTaskAnnotation}(\mathcal{T})$

\FOR{$i=1$ {\bfseries to} $n$}
    \IF{$\mathcal{T}$ has subtask}
        \STATE Identify subtask: $t_i \leftarrow \texttt{InferSubtask}(\text{SMILES}_i^{\text{raw}})$
        \STATE Load subtask annotations: $\mathcal{P}_{\text{subtask}} \leftarrow \texttt{LoadSubtaskAnnotation}(t_i)$
    \ELSE
        \STATE $\mathcal{P}_{\text{subtask}} \leftarrow \emptyset$
    \ENDIF

    \STATE Assemble full prompt: $\mathcal{P}_i \leftarrow \texttt{Assemble}(\mathcal{P}_{\text{base}}, \mathcal{P}_{\text{task}}, \mathcal{P}_{\text{subtask}}, \text{SMILES}_i^{\text{raw}}, \text{IMG}_i)$
    \STATE Inject reasoning chain: $\mathcal{P}_i \leftarrow \mathcal{P}_i + \texttt{InsertCoTReasoningSteps}()$
    \STATE Inject few-shot examples (optional): $\mathcal{P}_i \leftarrow \mathcal{P}_i + \texttt{InsertFewShotExamples}()$
    \STATE Append output constraints: $\mathcal{P}_i \leftarrow \mathcal{P}_i + \texttt{InsertOutputFormatSpec}()$
    \STATE Query MLLM: $\mathcal{S}_i^{\text{rep}} \leftarrow \mathcal{M}(\mathcal{P}_i)$
    \STATE Add to output: $\mathcal{S}^{\text{rep}} \leftarrow \mathcal{S}^{\text{rep}} \cup \mathcal{S}_i^{\text{rep}}$
\ENDFOR

\RETURN $\mathcal{S}^{\text{rep}}$
\end{algorithmic}
\end{algorithm}

Each complete prompt $\mathcal{P}_i$ is assembled by integrating molecular SMILES, 2D image features, and mechanism-specific constraints. The prompt is then augmented with optional Chain-of-Thought (CoT) reasoning steps and few-shot exemplars to enhance model interpretability and generalization. Finally, strict output format constraints are appended to ensure syntactic and chemical validity. The multimodal prompt $\mathcal{P}_i$ is passed to the MLLMs $\mathcal{M}$ to generate candidate repaired molecules $\mathcal{S}_i^{\text{rep}}$, which are collected across all samples to produce the final output set $\mathcal{S}^{\text{rep}}$.

\section{Supplementary Materials for ToxiEval}\label{ToxiEval_sm}

\subsection{Symbol Definitions for ToxiEval Evaluation Framework}\label{ToxiEval_symbol}

Table~\ref{tab:symbol_definitions_toxieval} defines the notations used in the ToxiEval evaluation framework described in Section~\ref{ToxiEval}. These symbols cover the key components of the multi-criteria decision process, including the structure of candidate sets, task-aware safety prediction, cheminformatics-derived molecular properties, and threshold-based decision variables.

\begin{table*}[!htbp]
 \small
\centering
\caption{\textbf{Symbol Definitions for the ToxiEval Evaluation Framework.}}
\label{tab:symbol_definitions_toxieval}
\begin{tabularx}{\textwidth}{@{}lX@{}}
\toprule
\textbf{Symbol} & \textbf{Definition} \\
\midrule
$\text{SMILES}_i^{\text{raw}}$ & Original toxicity-positive molecule requiring repair. \\
$\mathcal{S}_i^{\text{rep}} = \{ \text{SMILES}_i^{\text{rep}(1)}, \text{SMILES}_i^{\text{rep}(2)}, \text{SMILES}_i^{\text{rep}(3)} \}$ & Candidate repaired molecules generated by the MLLM for input $i$. \\
$s \in \mathcal{S}_i^{\text{rep}}$ & A single candidate molecule from the repaired set. \\
$\mathcal{T}$ & Toxicity repair task identifier (e.g., AMES, DILI, hERG, LD50). \\
$\mathscr{G}_{\text{tox}}$ & Toxicity prediction model used for safety scoring (e.g., TxGemma-Predict). \\
$\mathcal{S}_{\text{safe}}$ & Safety score of $s$ based on task $\mathcal{T}$ (binary or normalized). \\
$\mathcal{Q}$ & QED (Quantitative Estimate of Drug-likeness) score of candidate $s$, in $[0,1]$. \\
$\mathcal{S}_{\text{sas}}$ & Synthetic Accessibility Score of candidate $s$, in $[1,10]$ (lower is better). \\
$\mathcal{V}_{\text{ro5}}$ & Number of violations of Lipinski's Rule of Five by candidate $s$. \\
$\mathcal{S}_{\text{sim}}$ & Tanimoto similarity between candidate $s$ and original molecule $\text{SMILES}_i^{\text{raw}}$. \\
$\mathcal{C}^{\ast}$ & Set of evaluation thresholds for all criteria (e.g., $\mathcal{Q} \geq 0.5$, $\mathcal{S}_{\text{sas}} \leq 6$). \\
$\texttt{success}_i \in \{0,1\}$ & Boolean label indicating whether any $s \in \mathcal{S}_i^{\text{rep}}$ satisfies all criteria. \\
\bottomrule
\end{tabularx}
\end{table*}

\subsection{Intuitive Explanation of ToxiEval Evaluation Metrics}\label{appendix_metrics}

This section provides conceptual explanations for the five evaluation metrics used in the ToxiEval framework, helping readers understand their semantic meanings and practical relevance in the context of molecular toxicity repair.

\paragraph{Quantitative Estimate of Drug-likeness (QED, $\mathcal{Q}$).}
QED quantifies how closely a molecule resembles typical drug-like compounds. A candidate molecule must retain adequate drug-likeness even after successful reduction in toxicity to be considered pharmaceutically viable. The QED score ranges from $0$ to $1$, with higher values indicating better alignment with known drug-like properties. It integrates multiple molecular features, including Molecular Weight (MW), Partition coefficient (logP), and the number of Hydrogen Bond Donors (HBD) and Acceptors (HBA), among others. A significantly lower QED score in a repaired molecule $\mathcal{S}_i$ compared to its toxic precursor $\mathcal{S}_i^{\text{raw}}$ may suggest a loss of drug development potential, despite successful toxicity mitigation.

\paragraph{Lipinski's Rule of Five (RO5, $\mathcal{V}_{\text{ro5}}$).}
RO5 is a widely adopted heuristic for evaluating the oral bioavailability of small molecules. It defines four property-based constraints: MW $\leq$ 500 Da, logP $\leq$ 5, HBD $\leq$ 5, and HBA $\leq$ 10. A molecule that violates more than one of these conditions (i.e., $\mathcal{V}_{\text{ro5}} > 1$) is generally considered less likely to exhibit favorable oral absorption. Structural modifications during toxicity repair may increase molecular weight or lipophilicity, resulting in elevated RO5 violations. Although originally designed for oral drugs, RO5 remains a valuable indicator of drug-likeness and structural feasibility across broader chemical design spaces.

\paragraph{Synthetic Accessibility Score (SAS, $\mathcal{S}_{\text{sas}}$).}
SAS estimates how easily a molecule can be synthesized in a laboratory setting. Scores range from $1$ (very easy) to $10$ (very difficult). High scores reflect complex ring systems, rare substructures, or unstable fragments that hinder synthesis. In ToxiEval, we use $\mathcal{S}_{\text{sas}}$ to assess whether a generated candidate is practically accessible, thereby bridging the gap between virtual optimization and real-world chemical feasibility.

\paragraph{Structural Similarity (SS, $\mathcal{S}_{\text{sim}}$).}
To ensure that the semantic identity of a molecule is preserved during repair, ToxiEval computes the Tanimoto similarity between the repaired molecule and the original toxic precursor. Let $\mathbf{f}^{\text{raw}}$ and $\mathbf{f}^{(i)}$ denote the fingerprint vectors of $\mathcal{S}_i^{\text{raw}}$ and its repaired version $s_i$, respectively. The Tanimoto similarity is defined as:
\[
\mathcal{S}_{\text{sim}} = \frac{|\mathbf{f}^{\text{raw}} \cap \mathbf{f}^{(i)}|}{|\mathbf{f}^{\text{raw}} \cup \mathbf{f}^{(i)}|}.
\]
Here, Extended Connectivity Fingerprints (e.g., ECFP6) are typically used. Higher values of $\mathcal{S}_{\text{sim}}$ indicate that the core scaffold remains intact, implying a mild structural adjustment. Conversely, low similarity implies substantial structural changes that may compromise molecular identity or original pharmacophores.
\section{Success Rates of Each Model on the 12 Sub-Tasks of Tox21}\label{Tox21_success}
Table~\ref{tab:tox21_success} reports the success rates of representative MLLMs on the 12 Tox21 subtasks, where each subtask is evaluated independently under the same structure-level success criterion in \texttt{ToxiEval}. The subtask identifiers (T1--T12), their names, and associated toxicity mechanisms are summarized in Table~\ref{tab:tox21_subtasks}. Overall, performance exhibits pronounced heterogeneity across mechanisms. Across model families, nuclear receptor (NR) endpoints are generally easier: \textbf{T5} (ER\_LBD), \textbf{T7} (AR), and \textbf{T9} (ER) achieve consistently high success rates for most models, while \textbf{T3} (Aromatase) also remains moderately solvable. In contrast, several stress-response and genotoxicity-related endpoints are substantially more difficult, most notably \textbf{T1} (SR\_ARE) and \textbf{T2} (SR\_p53), which often remain at or near zero across many models, indicating persistent difficulty in producing repairs that satisfy these mechanism-specific constraints. Interestingly, \textbf{T8} (NR\_PPAR-$\gamma$) also shows relatively low success across models, suggesting that certain receptor-modulation subtasks remain challenging despite the generally strong performance on NR categories. Taken together, the subtask breakdown highlights that toxicity repair difficulty is mechanism-dependent rather than uniform, and that current MLLMs exhibit uneven generalization across distinct toxicological pathways.

\begin{table*}[htbp]
  \setlength{\tabcolsep}{8pt}
  \normalsize
  \centering
  \caption{\textbf{Success Rates of Each Model on the 12 Sub-Tasks of Tox21.}}
  \label{tab:tox21_success}
  \vspace{2mm}
  \renewcommand{\arraystretch}{0.95}
  \resizebox{\linewidth}{!}{
\begin{tabular}{@{}lcccccccccccc@{}}
\toprule
\textbf{Models} & \textbf{T1} & \textbf{T2} & \textbf{T3} & \textbf{T4} & \textbf{T5} & \textbf{T6} & \textbf{T7} & \textbf{T8} & \textbf{T9} & \textbf{T10} & \textbf{T11} & \textbf{T12} \\
\midrule
\multicolumn{13}{c}{\small{\textbf{Domain-Specific MLLMs}}} \\
\midrule
ChemVLM 26B~\cite{li2025chemvlm} & 0.0 & 0.0 & 40.0 & 0.0 & 80.0 & 20.0 & 100.0 & 0.0 & 80.0 & 0.0 & 0.0 & 60.0 \\
ChemVLM 8B~\cite{li2025chemvlm} & 0.0 & 0.0 & 40.0 & 0.0 & 100.0 & 40.0 & 100.0 & 0.0 & 60.0 & 20.0 & 20.0 & 80.0 \\
Intern-S1~\cite{bai2025intern} & 0.0 & 20.0 & 40.0 & 40.0 & 100.0 & 20.0 & 100.0 & 20.0 & 80.0 & 20.0 & 60.0 & 80.0 \\
Intern-S1 Mini~\cite{bai2025intern} & 0.0 & 0.0 & 60.0 & 20.0 & 100.0 & 40.0 & 100.0 & 20.0 & 80.0 & 40.0 & 40.0 & 40.0 \\
\midrule

\textbf{Average Success Rate} & 0.0 & 5.0 & 45.0 & 15.0 & 95.0 & 40.0 & 100.0 & 10.0 & 75.0 & 30.0 & 25.0 & 65.0 \\

\midrule
\multicolumn{13}{c}{\small{\textbf{Close Source MLLMs}}} \\
\midrule
GPT-5.2~\cite{openai_gpt52_2025} & 0.0 & 20.0 & 40.0 & 40.0 & 100.0 & 20.0 & 100.0 & 20.0 & 80.0 & 40.0 & 40.0 & 80.0 \\
GPT-5.1~\cite{openai_gpt51_2025} & 0.0 & 20.0 & 40.0 & 20.0 & 80.0 & 20.0 & 100.0 & 0.0 & 80.0 & 40.0 & 20.0 & 100.0 \\
GPT-4.1~\cite{openai2025gpt41} & 0.0 & 20.0 & 40.0 & 20.0 & 80.0 & 60.0 & 100.0 & 0.0 & 80.0 & 40.0 & 40.0 & 100.0 \\
GPT-o3~\cite{openai_o3_o4mini_2025} & 0.0 & 20.0 & 40.0 & 20.0 & 100.0 & 20.0 & 100.0 & 20.0 & 80.0 & 40.0 & 60.0 & 100.0 \\
GPT-4o~\cite{hurst2024gpt} & 0.0 & 20.0 & 40.0 & 20.0 & 100.0 & 20.0 & 100.0 & 20.0 & 100.0 & 40.0 & 40.0 & 100.0 \\
GPT-o4-mini~\cite{openai_o3_o4mini_2025} & 20.0 & 20.0 & 40.0 & 40.0 & 100.0 & 20.0 & 100.0 & 20.0 & 80.0 & 40.0 & 40.0 & 100.0 \\

Claude Opus 4.5~\cite{anthropic_claude_opus4_5_2025} & 20.0 & 0.0 & 60.0 & 20.0 & 100.0 & 40.0 & 100.0 & 20.0 & 100.0 & 40.0 & 40.0 & 80.0 \\
Claude Sonnet 4.5~\cite{anthropic_claude_sonnet4_5_2025} & 20.0 & 20.0 & 20.0 & 40.0 & 100.0 & 60.0 & 100.0 & 20.0 & 100.0 & 40.0 & 60.0 & 80.0 \\
Claude 3.7 Sonnet~\cite{claude37sonnet2024} & 20.0 & 20.0 & 40.0 & 20.0 & 100.0 & 40.0 & 100.0 & 20.0 & 80.0 & 40.0 & 40.0 & 80.0 \\

Gemini 3 Pro~\cite{google_gemini3_pro_official_2025} & 0.0 & 0.0 & 40.0 & 20.0 & 100.0 & 40.0 & 100.0 & 20.0 & 80.0 & 40.0 & 40.0 & 80.0 \\
Gemini 2.5 Pro~\cite{gemini25proexp2025} & 0.0 & 0.0 & 60.0 & 40.0 & 100.0 & 40.0 & 100.0 & 20.0 & 80.0 & 40.0 & 40.0 & 80.0 \\
Gemini 3 Flash~\cite{google_gemini3_pro_official_2025} & 0.0 & 20.0 & 40.0 & 20.0 & 100.0 & 40.0 & 100.0 & 20.0 & 80.0 & 40.0 & 40.0 & 80.0 \\
Gemini 2.5 Flash~\cite{google_gemini2_5_flash_2025} & 0.0 & 20.0 & 40.0 & 40.0 & 100.0 & 40.0 & 100.0 & 40.0 & 80.0 & 40.0 & 60.0 & 100.0 \\

Grok 4~\cite{xai_grok4_2025} & 0.0 & 20.0 & 40.0 & 20.0 & 100.0 & 60.0 & 80.0 & 20.0 & 80.0 & 40.0 & 20.0 & 80.0 \\
Grok 2 Vision~\cite{xai_grok2_vision_2024} & 20.0 & 20.0 & 20.0 & 0.0 & 100.0 & 40.0 & 80.0 & 20.0 & 80.0 & 40.0 & 60.0 & 80.0 \\

GLM-4V Plus~\cite{GLM4VPlus2024} & 0.0 & 0.0 & 40.0 & 20.0 & 80.0 & 20.0 & 100.0 & 0.0 & 80.0 & 20.0 & 0.0 & 80.0 \\

Doubao Seed 1.6 Vision~\cite{bytedance_doubao_seed1_6_vision_2024} & 0.0 & 0.0 & 40.0 & 0.0 & 100.0 & 20.0 & 60.0 & 0.0 & 80.0 & 40.0 & 0.0 & 80.0 \\
Doubao 1.5 Thinking-Vision Pro~\cite{volcengine_doubao_1_5_thinking_vision_pro_2025} & 0.0 & 0.0 & 40.0 & 0.0 & 80.0 & 20.0 & 100.0 & 0.0 & 80.0 & 40.0 & 20.0 & 80.0 \\

Moonshot v1 128K Vision~\cite{team2025kimi} & 0.0 & 20.0 & 40.0 & 20.0 & 100.0 & 80.0 & 80.0 & 20.0 & 80.0 & 0.0 & 60.0 & 80.0 \\
Hunyuan Vision~\cite{tencent2024hunyuanvision} & 0.0 & 0.0 & 20.0 & 40.0 & 100.0 & 40.0 & 100.0 & 0.0 & 40.0 & 40.0 & 0.0 & 60.0 \\

\midrule

\textbf{Average Success Rate} & 5.0 & 13.0 & 39.0 & 23.0 & 96.0 & 37.0 & 95.0 & 15.0 & 81.0 & 37.0 & 36.0 & 85.0 \\

\midrule
\multicolumn{13}{c}{\small{\textbf{Open Source MLLMs}}} \\
\midrule
\rowcolor{gray!15} \multicolumn{13}{l}{$\blacktriangledown$ \emph{Scale $<14$B}} \\
DeepSeek-VL2 Tiny~\cite{wu2024deepseek} & 0.0 & 0.0 & 40.0 & 0.0 & 80.0 & 40.0 & 100.0 & 20.0 & 80.0 & 0.0 & 20.0 & 60.0 \\
Qwen2.5-VL 3B Instruct~\cite{bai2025qwen2} & 0.0 & 0.0 & 20.0 & 0.0 & 60.0 & 0.0 & 40.0 & 0.0 & 20.0 & 20.0 & 20.0 & 40.0 \\
LLaVA-OneVision 7B~\cite{li2024llava} & 20.0 & 0.0 & 40.0 & 0.0 & 80.0 & 20.0 & 100.0 & 40.0 & 80.0 & 0.0 & 20.0 & 100.0 \\
MiMo v2 Flash~\cite{xiaomimimo_mimo_v2_flash_2025} & 0.0 & 20.0 & 40.0 & 20.0 & 100.0 & 60.0 & 80.0 & 20.0 & 80.0 & 0.0 & 60.0 & 80.0 \\
InternVL 3.5 8B~\cite{wang2025internvl3} & 0.0 & 20.0 & 20.0 & 40.0 & 100.0 & 20.0 & 100.0 & 20.0 & 80.0 & 40.0 & 40.0 & 80.0 \\
Qwen2.5-VL 7B Instruct~\cite{bai2025qwen2} & 0.0 & 0.0 & 0.0 & 0.0 & 20.0 & 0.0 & 0.0 & 0.0 & 0.0 & 0.0 & 0.0 & 0.0 \\

\rowcolor{gray!15} \multicolumn{13}{l}{$\blacktriangledown$ \emph{14B$\leq$Scale$<72$B}} \\
InternVL 3.5 14B~\cite{wang2025internvl3} & 0.0 & 20.0 & 20.0 & 20.0 & 100.0 & 20.0 & 100.0 & 20.0 & 80.0 & 20.0 & 40.0 & 80.0 \\
DeepSeek-VL2 Small~\cite{wu2024deepseek} & 0.0 & 0.0 & 40.0 & 20.0 & 100.0 & 40.0 & 80.0 & 20.0 & 80.0 & 0.0 & 20.0 & 80.0 \\
DeepSeek-VL2~\cite{wu2024deepseek} & 0.0 & 0.0 & 20.0 & 0.0 & 80.0 & 40.0 & 100.0 & 40.0 & 80.0 & 0.0 & 20.0 & 60.0 \\
Yi-Vision~\cite{ai2024yi} & 0.0 & 20.0 & 20.0 & 40.0 & 100.0 & 0.0 & 100.0 & 20.0 & 80.0 & 40.0 & 40.0 & 80.0 \\
InternVL 3.5 38B~\cite{wang2025internvl3} & 0.0 & 20.0 & 20.0 & 40.0 & 100.0 & 20.0 & 100.0 & 20.0 & 80.0 & 40.0 & 40.0 & 80.0 \\
Qwen2.5-VL 32B Instruct~\cite{bai2025qwen2} & 0.0 & 0.0 & 20.0 & 0.0 & 0.0 & 0.0 & 0.0 & 0.0 & 20.0 & 0.0 & 0.0 & 20.0 \\
Qwen3-VL 30B-A3B Instruct~\cite{bai2025qwen3vl} & 0.0 & 0.0 & 40.0 & 20.0 & 80.0 & 20.0 & 80.0 & 20.0 & 80.0 & 20.0 & 20.0 & 80.0 \\

\rowcolor{gray!15} \multicolumn{13}{l}{$\blacktriangledown$ \emph{Scale $\geq72$B}} \\
LLaVA-OneVision 72B~\cite{li2024llava} & 20.0 & 0.0 & 40.0 & 20.0 & 100.0 & 20.0 & 100.0 & 20.0 & 80.0 & 20.0 & 0.0 & 60.0 \\
Qwen2.5-VL 72B Instruct~\cite{bai2025qwen2} & 0.0 & 0.0 & 20.0 & 20.0 & 80.0 & 0.0 & 100.0 & 20.0 & 80.0 & 40.0 & 60.0 & 80.0 \\
InternVL 3 78B~\cite{zhu2025internvl3} & 0.0 & 20.0 & 40.0 & 0.0 & 100.0 & 60.0 & 80.0 & 20.0 & 80.0 & 40.0 & 40.0 & 80.0 \\
InternVL 3.5 241B-A28B~\cite{wang2025internvl3} & 0.0 & 20.0 & 60.0 & 60.0 & 100.0 & 40.0 & 100.0 & 20.0 & 80.0 & 60.0 & 60.0 & 80.0 \\
Qwen3-VL 235B-A22B Instruct~\cite{bai2025qwen3vl} & 0.0 & 20.0 & 40.0 & 20.0 & 80.0 & 40.0 & 100.0 & 20.0 & 80.0 & 40.0 & 40.0 & 100.0 \\
Qwen3-VL 235B-A22B Thinking~\cite{bai2025qwen3vl} & 0.0 & 0.0 & 40.0 & 40.0 & 0.0 & 20.0 & 100.0 & 0.0 & 80.0 & 60.0 & 40.0 & 100.0 \\

\midrule

\textbf{Average Success Rate} & 2.1 & 8.4 & 30.5 & 18.9 & 76.8 & 24.2 & 82.1 & 17.9 & 69.5 & 23.2 & 30.5 & 70.5 \\

\bottomrule
\end{tabular}}
\end{table*}
\section{Success Rates of Each Model on the 10 Selected Sub-Tasks of ToxCast}\label{Toxcast_success}

Table~\ref{tab:toxcast_success} reports success rates on 10 mechanism-diverse subtasks selected from ToxCast, with the corresponding identifiers and toxicity mechanisms summarized in Table~\ref{tab:toxcast_subtasks}. Compared to the Tox21 subtask suite, ToxCast exhibits a more uneven difficulty profile: several endpoints (e.g., \textbf{T1} mitochondrial stress, \textbf{T2} immune modulation, and \textbf{T9} GPCR-related signaling) achieve relatively high success across many models, whereas \textbf{T5} (ROR-$\gamma$ activation) and \textbf{T6} (p53 reporter activation) remain consistently low, indicating that certain receptor-activation and genotoxicity readouts pose stronger bottlenecks for structure-level repair. At the model level, closed-source MLLMs generally yield higher and more stable success rates than open-source counterparts, while domain-specific models (notably the \texttt{Intern-S1} family) are competitive on multiple subtasks. Overall, the ToxCast breakdown highlights substantial mechanism-dependent variability and underscores that robust toxicity repair requires generalization across heterogeneous biological pathways rather than a single toxicity phenotype.

\begin{table*}[!t]
  \setlength{\tabcolsep}{11pt}
  \normalsize
  \centering
  \caption{\textbf{Success Rates of Each Model on the 10 Selected Sub-Tasks of ToxCast.}}
  \label{tab:toxcast_success}
  \vspace{2mm}
  \renewcommand{\arraystretch}{0.95}
  \resizebox{\linewidth}{!}{
\begin{tabular}{@{}lcccccccccc@{}}
\toprule
\textbf{Models} & \textbf{T1} & \textbf{T2} & \textbf{T3} & \textbf{T4} & \textbf{T5} & \textbf{T6} & \textbf{T7} & \textbf{T8} & \textbf{T9} & \textbf{T10} \\
\midrule
\multicolumn{11}{c}{\small{\textbf{Domain-Specific MLLMs}}} \\
\midrule
ChemVLM 26B~\cite{li2025chemvlm} & 66.7 & 33.3 & 50.0 & 16.7 & 0.0 & 16.7 & 33.3 & 16.7 & 16.7 & 33.3 \\
ChemVLM 8B~\cite{li2025chemvlm} & 83.3 & 50.0 & 50.0 & 50.0 & 16.7 & 16.7 & 33.3 & 33.3 & 66.7 & 33.3 \\
Intern-S1~\cite{bai2025intern} & 83.3 & 66.7 & 66.7 & 50.0 & 16.7 & 16.7 & 66.7 & 66.7 & 66.7 & 50.0 \\
Intern-S1 Mini~\cite{bai2025intern} & 66.7 & 83.3 & 66.7 & 50.0 & 16.7 & 16.7 & 66.7 & 66.7 & 66.7 & 50.0 \\
\midrule

\textbf{Average Success Rate} & 75.0 & 58.3 & 54.2 & 41.7 & 12.5 & 16.7 & 41.7 & 37.5 & 58.3 & 41.7 \\

\midrule
\multicolumn{11}{c}{\small{\textbf{Close Source MLLMs}}} \\
\midrule
GPT-5.2~\cite{openai_gpt52_2025} & 83.3 & 66.7 & 50.0 & 66.7 & 16.7 & 33.3 & 50.0 & 50.0 & 66.7 & 50.0 \\
GPT-5.1~\cite{openai_gpt51_2025} & 83.3 & 66.7 & 66.7 & 66.7 & 16.7 & 16.7 & 66.7 & 66.7 & 66.7 & 33.3 \\
GPT-4.1~\cite{openai2025gpt41} & 83.3 & 66.7 & 50.0 & 33.3 & 16.7 & 16.7 & 50.0 & 50.0 & 66.7 & 33.3 \\
GPT-o3~\cite{openai_o3_o4mini_2025} & 100.0 & 83.3 & 50.0 & 50.0 & 16.7 & 16.7 & 66.7 & 66.7 & 83.3 & 33.3 \\
GPT-4o~\cite{hurst2024gpt} & 83.3 & 66.7 & 50.0 & 66.7 & 16.7 & 16.7 & 66.7 & 66.7 & 66.7 & 33.3 \\
GPT-o4-mini~\cite{openai_o3_o4mini_2025} & 83.3 & 66.7 & 50.0 & 50.0 & 33.3 & 16.7 & 66.7 & 33.3 & 83.3 & 33.3 \\

Claude Opus 4.5~\cite{anthropic_claude_opus4_5_2025} & 83.3 & 66.7 & 66.7 & 50.0 & 16.7 & 33.3 & 83.3 & 66.7 & 66.7 & 33.3 \\
Claude Sonnet 4.5~\cite{anthropic_claude_sonnet4_5_2025} & 83.3 & 66.7 & 66.7 & 66.7 & 0.0 & 16.7 & 83.3 & 66.7 & 83.3 & 33.3 \\
Claude 3.7 Sonnet~\cite{claude37sonnet2024} & 50.0 & 50.0 & 50.0 & 66.7 & 16.7 & 33.3 & 66.7 & 66.7 & 83.3 & 50.0 \\

Gemini 3 Pro~\cite{google_gemini3_pro_official_2025} & 83.3 & 66.7 & 50.0 & 33.3 & 16.7 & 33.3 & 50.0 & 50.0 & 83.3 & 33.3 \\
Gemini 2.5 Pro~\cite{gemini25proexp2025} & 100.0 & 66.7 & 50.0 & 33.3 & 33.3 & 33.3 & 66.7 & 33.3 & 83.3 & 50.0 \\
Gemini 3 Flash~\cite{google_gemini3_pro_official_2025} & 83.3 & 50.0 & 50.0 & 33.3 & 16.7 & 33.3 & 66.7 & 33.3 & 83.3 & 33.3 \\
Gemini 2.5 Flash~\cite{google_gemini2_5_flash_2025} & 83.3 & 83.3 & 66.7 & 50.0 & 33.3 & 33.3 & 83.3 & 50.0 & 83.3 & 50.0 \\

Grok 4~\cite{xai_grok4_2025} & 83.3 & 50.0 & 50.0 & 50.0 & 0.0 & 16.7 & 33.3 & 50.0 & 66.7 & 33.3 \\
Grok 2 Vision~\cite{xai_grok2_vision_2024} & 83.3 & 66.7 & 50.0 & 50.0 & 0.0 & 16.7 & 33.3 & 50.0 & 66.7 & 33.3 \\

GLM-4V Plus~\cite{GLM4VPlus2024} & 83.3 & 50.0 & 50.0 & 50.0 & 0.0 & 16.7 & 33.3 & 33.3 & 83.3 & 50.0 \\

Doubao Seed 1.6 Vision~\cite{bytedance_doubao_seed1_6_vision_2024} & 66.7 & 66.7 & 33.3 & 50.0 & 16.7 & 16.7 & 66.7 & 16.7 & 66.7 & 33.3 \\
Doubao 1.5 Thinking-Vision Pro~\cite{volcengine_doubao_1_5_thinking_vision_pro_2025} & 83.3 & 66.7 & 50.0 & 66.7 & 0.0 & 16.7 & 50.0 & 66.7 & 83.3 & 33.3 \\

Moonshot v1 128K Vision~\cite{team2025kimi} & 50.0 & 33.3 & 50.0 & 50.0 & 16.7 & 33.3 & 50.0 & 50.0 & 33.3 & 33.3 \\
Hunyuan Vision~\cite{tencent2024hunyuanvision} & 50.0 & 50.0 & 50.0 & 33.3 & 0.0 & 16.7 & 33.3 & 33.3 & 33.3 & 16.7 \\
\midrule

\textbf{Average Success Rate} & 79.1 & 62.5 & 52.5 & 52.5 & 15.8 & 23.3 & 59.2 & 51.7 & 71.7 & 35.0 \\

\midrule
\multicolumn{11}{c}{\small{\textbf{Open Source MLLMs}}} \\
\midrule
\rowcolor{gray!15} \multicolumn{11}{l}{$\blacktriangledown$ \emph{Scale $<14$B}} \\
DeepSeek-VL2 Tiny~\cite{wu2024deepseek} & 50.0 & 50.0 & 50.0 & 16.7 & 0.0 & 16.7 & 33.3 & 33.3 & 50.0 & 33.3 \\
Qwen2.5-VL 3B Instruct~\cite{bai2025qwen2} & 66.7 & 50.0 & 66.7 & 33.3 & 0.0 & 0.0 & 50.0 & 33.3 & 66.7 & 33.3 \\
LLaVA-OneVision 7B~\cite{li2024llava} & 83.3 & 33.3 & 50.0 & 33.3 & 0.0 & 16.7 & 33.3 & 16.7 & 50.0 & 16.7 \\
MiMo v2 Flash~\cite{xiaomimimo_mimo_v2_flash_2025} & 83.3 & 66.7 & 50.0 & 50.0 & 16.7 & 33.3 & 16.7 & 66.7 & 100.0 & 33.3 \\
InternVL 3.5 8B~\cite{wang2025internvl3} & 83.3 & 50.0 & 66.7 & 50.0 & 50.0 & 33.3 & 33.3 & 50.0 & 100.0 & 50.0 \\
Qwen2.5-VL 7B Instruct~\cite{bai2025qwen2} & 0.0 & 33.3 & 16.7 & 0.0 & 0.0 & 0.0 & 33.3 & 0.0 & 0.0 & 0.0 \\

\rowcolor{gray!15} \multicolumn{11}{l}{$\blacktriangledown$ \emph{14B$\leq$Scale$<72$B}} \\
InternVL 3.5 14B~\cite{wang2025internvl3} & 83.3 & 66.7 & 66.7 & 66.7 & 0.0 & 33.3 & 33.3 & 50.0 & 83.3 & 50.0 \\
DeepSeek-VL2 Small~\cite{wu2024deepseek} & 83.3 & 50.0 & 66.7 & 50.0 & 0.0 & 16.7 & 33.3 & 33.3 & 50.0 & 33.3 \\
DeepSeek-VL2~\cite{wu2024deepseek} & 83.3 & 33.3 & 33.3 & 66.7 & 0.0 & 16.7 & 33.3 & 33.3 & 66.7 & 33.3 \\
Yi-Vision~\cite{ai2024yi} & 83.3 & 50.0 & 50.0 & 33.3 & 16.7 & 33.3 & 33.3 & 66.7 & 83.3 & 33.3 \\
InternVL 3.5 38B~\cite{wang2025internvl3} & 100.0 & 66.7 & 50.0 & 66.7 & 50.0 & 16.7 & 50.0 & 50.0 & 66.7 & 50.0 \\
Qwen2.5-VL 32B Instruct~\cite{bai2025qwen2} & 16.7 & 16.7 & 16.7 & 0.0 & 0.0 & 0.0 & 16.7 & 0.0 & 16.7 & 0.0 \\
Qwen3-VL 30B-A3B Instruct~\cite{bai2025qwen3vl} & 66.7 & 33.3 & 50.0 & 16.7 & 16.7 & 16.7 & 16.7 & 33.3 & 50.0 & 33.3 \\

\rowcolor{gray!15} \multicolumn{11}{l}{$\blacktriangledown$ \emph{Scale $\geq72$B}} \\
LLaVA-OneVision 72B~\cite{li2024llava} & 66.7 & 50.0 & 50.0 & 66.7 & 0.0 & 16.7 & 0.0 & 33.3 & 33.3 & 33.3 \\
Qwen2.5-VL 72B Instruct~\cite{bai2025qwen2} & 66.7 & 66.7 & 33.3 & 50.0 & 0.0 & 33.3 & 50.0 & 66.7 & 83.3 & 33.3 \\
InternVL 3 78B~\cite{zhu2025internvl3} & 66.7 & 50.0 & 33.3 & 50.0 & 16.7 & 16.7 & 50.0 & 50.0 & 83.3 & 33.3 \\
InternVL 3.5 241B-A28B~\cite{wang2025internvl3} & 83.3 & 83.3 & 66.7 & 66.7 & 33.3 & 16.7 & 66.7 & 66.7 & 83.3 & 50.0 \\
Qwen3-VL 235B-A22B Instruct~\cite{bai2025qwen3vl} & 83.3 & 66.7 & 33.3 & 50.0 & 16.7 & 16.7 & 33.3 & 50.0 & 66.7 & 33.3 \\
Qwen3-VL 235B-A22B Thinking~\cite{bai2025qwen3vl} & 83.3 & 66.7 & 50.0 & 50.0 & 33.3 & 16.7 & 66.7 & 50.0 & 66.7 & 33.3 \\
\midrule

\textbf{Average Success Rate} & 70.2 & 51.8 & 47.4 & 43.0 & 13.2 & 18.4 & 36.0 & 41.2 & 63.2 & 32.4 \\

\bottomrule
\end{tabular}}
\label{table:main-result}
\end{table*}
\section{Evaluation of Molecular Structure Validity}\label{Valid}

\begin{table*}[!t]
  \setlength{\tabcolsep}{3pt}
  \normalsize
  \centering
  \caption{\textbf{Toxicity Repair Valid Rates (\%).} hERG$\text{C}$, hERG$\text{K}$, and Carc$_\text{L}$ refer to the \texttt{hERG\_Central}, \texttt{hERG\_Karim}, and \texttt{Carcinogens} tasks, respectively. \texttt{SkinRxn} denotes the \texttt{SkinReaction} task. }
  \label{tab:MLLM_valid}
  \vspace{2mm}
  \renewcommand{\arraystretch}{0.95}
  \resizebox{\linewidth}{!}{
\begin{tabular}{@{}lcccccccccccc@{}}
\toprule
\textbf{Models} & \textbf{LD50} & \textbf{hERG$_C$} & \textbf{Carc$_L$} & \textbf{Tox21} & \textbf{AMES} & \textbf{ToxCast} & \textbf{ClinTox} & \textbf{DILI} & \textbf{hERG$_K$} & \textbf{hERG} & \textbf{SkinRxn} & \textbf{Overall} \\
\midrule
\multicolumn{13}{c}{\small{\textbf{Domain-Specific MLLMs}}} \\
\midrule
ChemVLM 26B~\cite{li2025chemvlm} & 98.3 & 55.0 & 31.7 & 68.3 & 11.7 & 66.7 & 71.7 & 18.3 & 25.0 & 20.0 & 20.0 & 44.2 \\
ChemVLM 8B~\cite{li2025chemvlm} & 100.0 & 45.0 & 38.3 & 71.7 & 25.0 & 83.3 & 66.7 & 11.7 & 21.7 & 18.3 & 21.7 & 45.8 \\
Intern-S1~\cite{bai2025intern} & 100.0 & 80.0 & 66.7 & 80.0 & 43.3 & 85.0 & 80.0 & 11.7 & 51.7 & 26.7 & 43.3 & 60.8 \\
Intern-S1 Mini~\cite{bai2025intern} & 95.0 & 78.3 & 65.0 & 71.7 & 38.3 & 83.3 & 81.7 & 11.7 & 38.3 & 31.7 & 36.7 & 57.4 \\
\midrule

\textbf{Average Success Rate} & 98.3 & 64.6 & 50.4 & 72.9 & 29.6 & 79.6 & 75.0 & 13.3 & 34.2 & 24.2 & 30.4 & 52.0 \\

\midrule
\multicolumn{13}{c}{\small{\textbf{Close Source MLLMs}}} \\
\midrule
GPT-5.2~\cite{openai_gpt52_2025} & 100.0 & 83.3 & 61.7 & 83.3 & 41.7 & 86.7 & 78.3 & 8.3 & 68.3 & 36.7 & 43.3 & 62.9 \\
GPT-5.1~\cite{openai_gpt51_2025} & 78.3 & 86.7 & 21.7 & 71.7 & 73.3 & 100.0 & 65.0 & 46.7 & 46.7 & 10.0 & 43.3 & 58.5 \\
GPT-4.1~\cite{openai2025gpt41} & 100 & 73.3 & 68.3 & 78.3 & 35.0 & 85.0 & 75.0 & 10.0 & 40.0 & 20.0 & 38.3 & 56.7 \\
GPT-o3~\cite{openai_o3_o4mini_2025} & 100.0 & 88.3 & 73.3 & 78.3 & 63.3 & 86.7 & 85.0 & 11.7 & 71.7 & 35.0 & 51.7 & 67.7 \\
GPT-4o~\cite{hurst2024gpt} & 100.0 & 68.3 & 55.0 & 75.0 & 41.7 & 85.0 & 70.0 & 11.7 & 36.7 & 13.3 & 35.0 & 53.8 \\
GPT-o4-mini~\cite{openai_o3_o4mini_2025} & 100.0 & 80.0 & 75.0 & 83.3 & 58.3 & 91.7 & 71.7 & 11.7 & 61.7 & 38.3 & 53.3 & 65.9 \\
Claude Opus 4.5~\cite{anthropic_claude_opus4_5_2025} & 100.0 & 98.3 & 83.3 & 81.7 & 66.7 & 86.7 & 81.7 & 15.0 & 76.7 & 43.3 & 53.3 & 71.5 \\
Claude Sonnet 4.5~\cite{anthropic_claude_sonnet4_5_2025} & 100.0 & 88.3 & 78.3 & 88.3 & 55.0 & 81.7 & 75.0 & 11.7 & 65.0 & 41.7 & 60.0 & 67.7 \\
Claude 3.7 Sonnet~\cite{claude37sonnet2024} & 100.0 & 90.0 & 75.0 & 83.3 & 55.0 & 88.3 & 75.0 & 18.3 & 65.0 & 36.7 & 55.0 & 67.4 \\
Gemini 3 Pro~\cite{google_gemini3_pro_official_2025} & 100.0 & 80.0 & 75.0 & 78.3 & 60.0 & 91.7 & 75.0 & 13.3 & 51.7 & 20.0 & 48.3 & 63.0 \\
Gemini 2.5 Pro~\cite{gemini25proexp2025} & 100.0 & 96.7 & 75.0 & 88.3 & 66.7 & 91.7 & 90.0 & 13.3 & 85.0 & 50.0 & 48.3 & 73.2 \\
Gemini 3 Flash~\cite{google_gemini3_pro_official_2025} & 100.0 & 85.0 & 56.7 & 78.3 & 61.7 & 90.0 & 83.3 & 15.0 & 60.0 & 20.0 & 41.7 & 62.9 \\
Gemini 2.5 Flash~\cite{google_gemini2_5_flash_2025} & 100.0 & 93.3 & 70.0 & 83.3 & 60.0 & 88.3 & 81.7 & 13.3 & 71.7 & 36.7 & 50.0 & 68.0 \\
Grok 4~\cite{xai_grok4_2025} & 100.0 & 91.7 & 61.7 & 76.7 & 53.3 & 85.0 & 76.7 & 11.7 & 66.7 & 31.7 & 48.3 & 63.9 \\
Grok 2 Vision~\cite{xai_grok2_vision_2024} & 100.0 & 68.3 & 58.3 & 73.3 & 35.0 & 81.7 & 71.7 & 10.0 & 31.7 & 30.0 & 46.7 & 55.1 \\
GLM-4V Plus~\cite{GLM4VPlus2024} & 88.3 & 36.7 & 38.3 & 66.7 & 21.7 & 80.0 & 61.7 & 13.3 & 20.0 & 11.7 & 21.7 & 41.8 \\
Doubao Seed 1.6 Vision~\cite{bytedance_doubao_seed1_6_vision_2024} & 90.0 & 65.0 & 40.0 & 63.3 & 33.3 & 75.0 & 61.7 & 10.0 & 31.7 & 18.3 & 33.3 & 47.4 \\
Doubao 1.5 Thinking-Vision Pro~\cite{volcengine_doubao_1_5_thinking_vision_pro_2025} & 96.7 & 68.3 & 56.7 & 75.0 & 35.0 & 80.0 & 73.3 & 8.3 & 41.7 & 16.7 & 40.0 & 53.8 \\
Moonshot v1 128K Vision~\cite{team2025kimi} & 95.0 & 65.0 & 60.0 & 80.0 & 45.0 & 88.3 & 70.0 & 11.7 & 43.3 & 21.7 & 36.7 & 56.1 \\
Hunyuan Vision~\cite{tencent2024hunyuanvision} & 80.0 & 43.3 & 40.0 & 63.3 & 20.0 & 63.3 & 65.0 & 8.3 & 30.0 & 10.0 & 23.3 & 40.6 \\
\midrule

\textbf{Average Success Rate} & 96.4 & 77.5 & 61.2 & 77.5 & 49.1 & 85.3 & 74.3 & 13.7 & 53.3 & 27.1 & 43.6 & 59.9 \\

\midrule
\multicolumn{13}{c}{\small{\textbf{Open Source MLLMs}}} \\
\midrule
\rowcolor{gray!15} \multicolumn{13}{l}{$\blacktriangledown$ \emph{Scale $<14$B}} \\
DeepSeek-VL2 Tiny~\cite{wu2024deepseek} & 93.3 & 36.7 & 28.3 & 70.0 & 6.7 & 70.0 & 60.0 & 8.3 & 13.3 & 15.0 & 13.3 & 37.7 \\

Qwen2.5-VL 3B Instruct~\cite{bai2025qwen2} & 93.3 & 48.3 & 35.0 & 55.0 & 25.0 & 76.7 & 60.0 & 8.3 & 26.7 & 20.0 & 20.0 & 42.6 \\
LLaVA-OneVision 7B~\cite{li2024llava} & 95.0 & 38.3 & 31.7 & 70.0 & 5.0 & 75.0 & 68.3 & 10.0 & 21.7 & 15.0 & 25.0 & 41.4 \\
MiMo v2 Flash~\cite{xiaomimimo_mimo_v2_flash_2025} & 100.0 & 70.0 & 50.0 & 81.7 & 36.7 & 81.7 & 70.0 & 13.3 & 50.0 & 25.0 & 40.0 & 56.2 \\
InternVL 3.5 8B~\cite{wang2025internvl3} & 100.0 & 71.7 & 61.7 & 75.0 & 33.3 & 83.3 & 75.0 & 10.0 & 31.7 & 21.7 & 41.7 & 55.0 \\
Qwen2.5-VL 7B Instruct~\cite{bai2025qwen2} & 86.7 & 73.3 & 55.0 & 75.0 & 61.7 & 71.7 & 70.0 & 46.7 & 71.7 & 36.7 & 41.7 & 62.7 \\

\rowcolor{gray!15} \multicolumn{13}{l}{$\blacktriangledown$ \emph{14B$\leq$Scale$<72$B}} \\
InternVL 3.5 14B~\cite{wang2025internvl3} & 100.0 & 78.3 & 60.0 & 75.0 & 36.7 & 85.0 & 76.7 & 16.7 & 45.0 & 26.7 & 31.7 & 57.4 \\
DeepSeek-VL2 Small~\cite{wu2024deepseek} & 98.3 & 38.3 & 35.0 & 71.7 & 8.3 & 76.7 & 65.0 & 11.7 & 21.7 & 10.0 & 20.0 & 41.5 \\
DeepSeek-VL2~\cite{wu2024deepseek} & 93.3 & 35.0 & 25.0 & 68.3 & 10.0 & 70.0 & 63.3 & 10.0 & 13.3 & 10.0 & 15.0 & 37.6 \\
Yi-Vision~\cite{ai2024yi} & 100.0 & 56.7 & 51.7 & 78.3 & 50.0 & 83.3 & 70.0 & 10.0 & 33.3 & 21.7 & 50.0 & 55.0 \\
InternVL 3.5 38B~\cite{wang2025internvl3} & 100.0 & 63.3 & 53.3 & 81.7 & 38.3 & 86.7 & 68.3 & 8.3 & 43.3 & 20.0 & 45.0 & 55.3 \\
Qwen2.5-VL 32B Instruct~\cite{bai2025qwen2} & 88.3 & 48.3 & 35.0 & 55.0 & 25.0 & 76.7 & 60.0 & 8.3 & 26.7 & 20.0 & 20.0 & 42.6 \\
Qwen3-VL 30B-A3B Instruct~\cite{bai2025qwen3vl} & 70.0 & 80.0 & 13.3 & 41.7 & 73.3 & 100.0 & 48.3 & 21.7 & 25.0 & 10.0 & 25.0 & 46.2 \\

\rowcolor{gray!15} \multicolumn{13}{l}{$\blacktriangledown$ \emph{Scale $\geq72$B}} \\
LLaVA-OneVision 72B~\cite{li2024llava} & 95.0 & 38.3 & 31.7 & 70.0 & 5.0 & 75.0 & 68.3 & 10.0 & 21.7 & 15.0 & 25.0 & 41.4 \\
Qwen2.5-VL 72B Instruct~\cite{bai2025qwen2} & 98.3 & 50.0 & 51.7 & 73.3 & 38.3 & 83.3 & 61.7 & 8.3 & 36.7 & 13.3 & 43.3 & 50.8 \\
InternVL 3 78B~\cite{zhu2025internvl3} & 98.3 & 55.0 & 46.7 & 71.7 & 26.7 & 81.7 & 65.0 & 11.7 & 28.3 & 16.7 & 36.7 & 48.9 \\
InternVL 3.5 241B-A28B~\cite{wang2025internvl3} & 100.0 & 75.0 & 61.7 & 80.0 & 51.7 & 85.0 & 75.0 & 11.7 & 58.3 & 33.3 & 38.3 & 60.9 \\
Qwen3-VL 235B-A22B Instruct~\cite{bai2025qwen3vl} & 75.0 & 80.0 & 16.7 & 66.7 & 80.0 & 100.0 & 53.3 & 33.3 & 38.3 & 10.0 & 41.7 & 54.1 \\
Qwen3-VL 235B-A22B Thinking~\cite{bai2025qwen3vl} & 98.3 & 61.7 & 73.3 & 68.3 & 45.0 & 86.7 & 78.3 & 10.0 & 46.7 & 16.7 & 50.0 & 57.7 \\

\midrule

\textbf{Average Success Rate} & 93.8 & 57.8 & 43.0 & 69.9 & 34.6 & 81.5 & 66.1 & 14.1 & 34.4 & 18.8 & 32.8 & 49.7 \\

\bottomrule
\end{tabular}}
\end{table*}

As shown in Table~\ref{tab:MLLM_valid}, structural validity varies substantially across both models and tasks. Overall, closed-source MLLMs achieve the highest average validity (Overall: 59.9\%), with several leading models exceeding 65\% (e.g., \texttt{Claude Opus 4.5} at 71.5\% and \texttt{Gemini 2.5 Pro} at 73.2\%). In comparison, open-source MLLMs exhibit a lower average validity (49.7\%), though the best open-source models still reach competitive validity levels (e.g., \texttt{Qwen2.5-VL 7B Instruct} at 62.7\% and \texttt{InternVL 3.5 241B-A28B} at~60.9\%). 

At the task level, validity is highly uneven. \texttt{LD50} consistently exhibits near-perfect validity for most models, indicating that generating syntactically parsable candidates is generally feasible under this endpoint. In contrast, \texttt{DILI} presents a pronounced validity bottleneck across nearly all model families (often near 10\%--20\% overall, even for strong closed-source models), implying that this task is particularly prone to invalid generations. Moreover, scaling trends are not strictly monotonic: larger models often improve validity (e.g., within \texttt{InternVL 3.5}, 241B-A28B exceeds 38B), yet exceptions exist (e.g., \texttt{Qwen3-VL 30B-A3B} exhibits low validity on \texttt{Carc$_L$} and modest overall validity despite strong performance elsewhere). Taken together, these results indicate that structure validity is a non-trivial bottleneck for certain endpoints and model families, motivating its role as a mandatory pre-filter in \texttt{ToxiEval} before applying toxicity and drug-likeness constraints.
\section{Additional Discussions and Outlook}\label{findings_appendix}

Our benchmark aggregates 11 toxicity prediction tasks from TDC~\cite{velez2024tdc, huang2021therapeutics, huang2022artificial}, each originating from distinct experimental settings and assay protocols. Following TDC's standard practice, we treat each dataset as an independent task-level abstraction rather than assuming label consistency across tasks. Toxicity prediction labels are evaluated using predefined prompts from TxGemma-Predict~\cite{wang2025txgemma}, thereby avoiding cross-task label mixing and mitigating potential inconsistencies introduced by experimental heterogeneity.

Within our current definition of the main task taxonomy, Nuclear Receptor (NR) related sub-tasks from Tox21 (such as NR\_AR\_LBD, NR\_ER, NR\_ER\_LBD, etc.)~\cite{huang2016tox21challenge}, as well as sub-tasks from ToxCast including NVS\_GPCR\_gLTB4, NVS\_ENZ\_hAChE, ATG\_ROR $\gamma$\_TRANS\_up~\cite{richard2016toxcast}, exhibit clear receptor- or target-specific orientations. In addition, the hERG~\cite{wang2016admet, karim2021cardiotox, du2011hergcentral} ion channel blockade endpoint is inherently driven by molecular interactions with protein targets. It should be emphasized that this benchmark is not designed for real receptor binding optimization problems and therefore does not rely on three-dimensional receptor structure information. Under the TDC~\cite{velez2024tdc, huang2021therapeutics, huang2022artificial} framework, the aforementioned endpoints are supervised by phenotypic labels derived from in vitro high-throughput screening (HTS) or related bioassays; receptor structures are not provided in the datasets, and a large number of classical baselines (such as those used in the Tox21 Challenge framework~\cite{huang2016tox21challenge}, DeepTox~\cite{mayr2016deeptox}) also rely solely on molecular structure representations for prediction. Under this setting, the performance of MLLMs should not be interpreted as explicitly modeling physical molecule–receptor interactions, but rather as arising from implicit use of structure–toxicity statistical associations and pattern generalization.

We acknowledge that the data used to evaluate MLLMs may have been encountered during model training, and thus may pose potential data leakage risks. This risk constitutes a structural limitation that is broadly present in current LLM/MLLM benchmarking and is difficult to fully eliminate. Therefore, the results presented in this work should be interpreted as an evaluation of the models' ability to invoke and integrate existing chemical knowledge, rather than as a strict assessment of generalization to entirely unseen~compounds.

\section{License and Usage Information for External Assets}\label{appendix_licenses}


\paragraph{RDKit~\cite{rdkit}.} Our framework extensively uses RDKit (version 2023.9.6) for various molecular operations, including QED calculation, Lipinski's Rule of Five evaluation (RO5), and molecular similarity computation. RDKit is an open-source cheminformatics software package released under the BSD 3-Clause "New" or "Revised" License. The official repository is available at \url{https://github.com/rdkit/rdkit}, and more information can be found at \url{https://www.rdkit.org}.

\paragraph{Synthetic Accessibility Score (SAS)~\cite{ertl2009estimation}.} The implementation of the Synthetic Accessibility Score in our framework is based on the work by Ertl and Schuffenhauer, as described in their paper "Estimation of Synthetic Accessibility Score of Drug-like Molecules based on Molecular Complexity and Fragment Contributions" (Journal of Cheminformatics 1:8, 2009). The implementation is adapted from the original code provided by the authors, which has been made publicly available. The SAS implementation is released with the same BSD 3-Clause License as RDKit, as it's typically distributed as part of the RDKit control package. The original paper is available at \url{https://doi.org/10.1186/1758-2946-1-8}.

\paragraph{TxGemma~\cite{wang2025txgemma}.} Our framework uses the TxGemma-27B-predict model from Google for toxicity prediction, accessible via the Hugging Face Transformers library. Unlike standard open-source models, TxGemma is governed by the Health AI Developer Foundations Terms of Use rather than the Apache 2.0 license. The model can be accessed at \url{https://huggingface.co/google/txgemma-27b-predict}, with usage terms available at \url{https://developers.google.com/health-ai-developer-foundations/terms}. The official GitHub repository provides additional implementation details and usage examples: \url{https://github.com/google-gemini/gemma-cookbook/tree/main/TxGemma}. A full model description is available in "TxGemma: Efficient and Agentic LLMs for Therapeutics" by Wang \emph{et al}. (2024), at \url{https://arxiv.org/abs/2504.06196}.

All external assets were used in accordance with their respective licenses and terms of use.
\section{Success and Failure Examples of Toxicity Repair}\label{successful_cases}

\begin{figure}[htbp]
  \centering
  \includegraphics[width=0.9\linewidth]{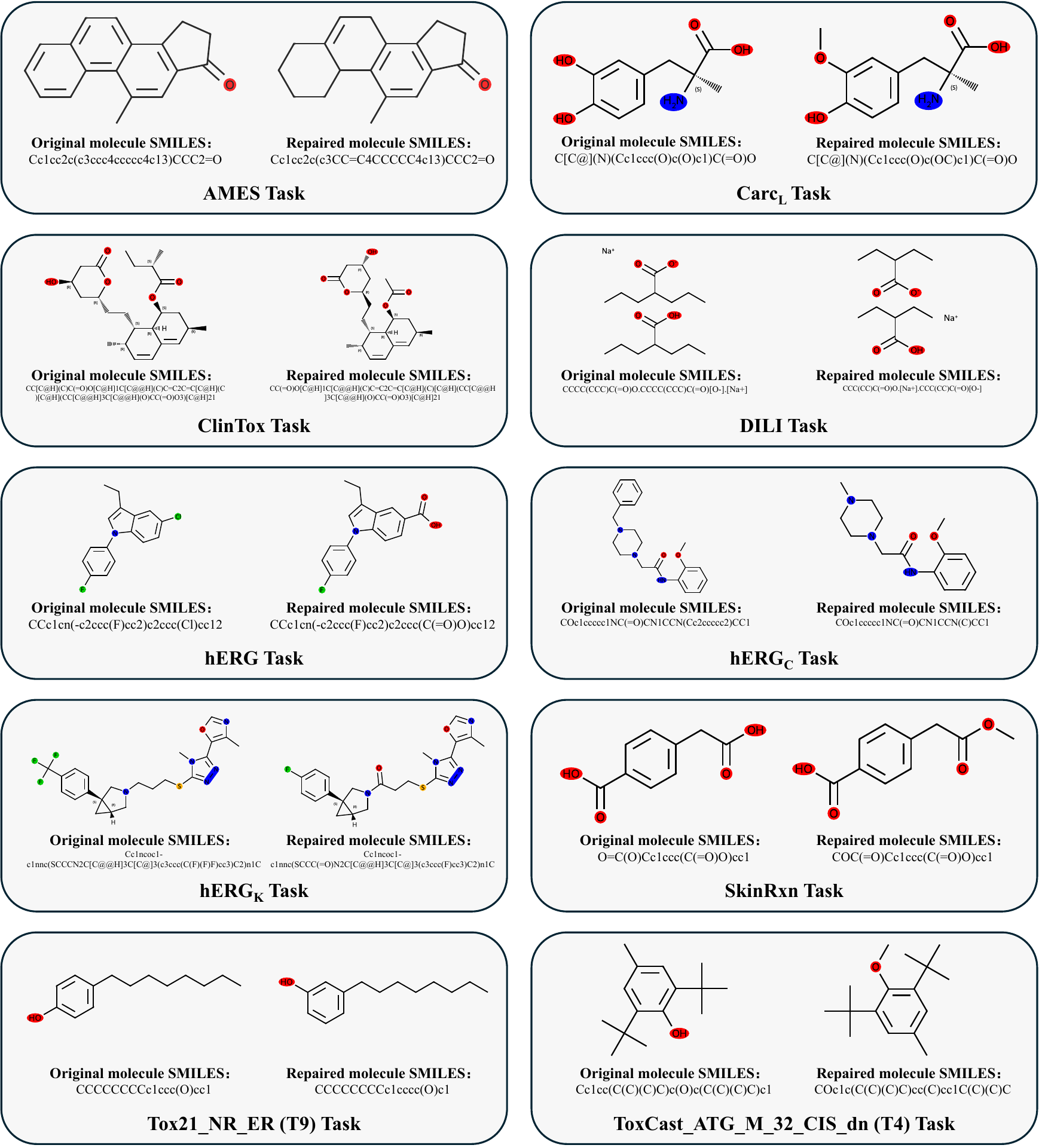}
  \caption{Representative successful toxicity repair cases of \texttt{Claude Opus 4.5}~\cite{anthropic_claude_opus4_5_2025}.}
  \label{fig:success_cases}
\end{figure}

\begin{figure}[htbp]
  \centering
  \includegraphics[width=0.9\linewidth]{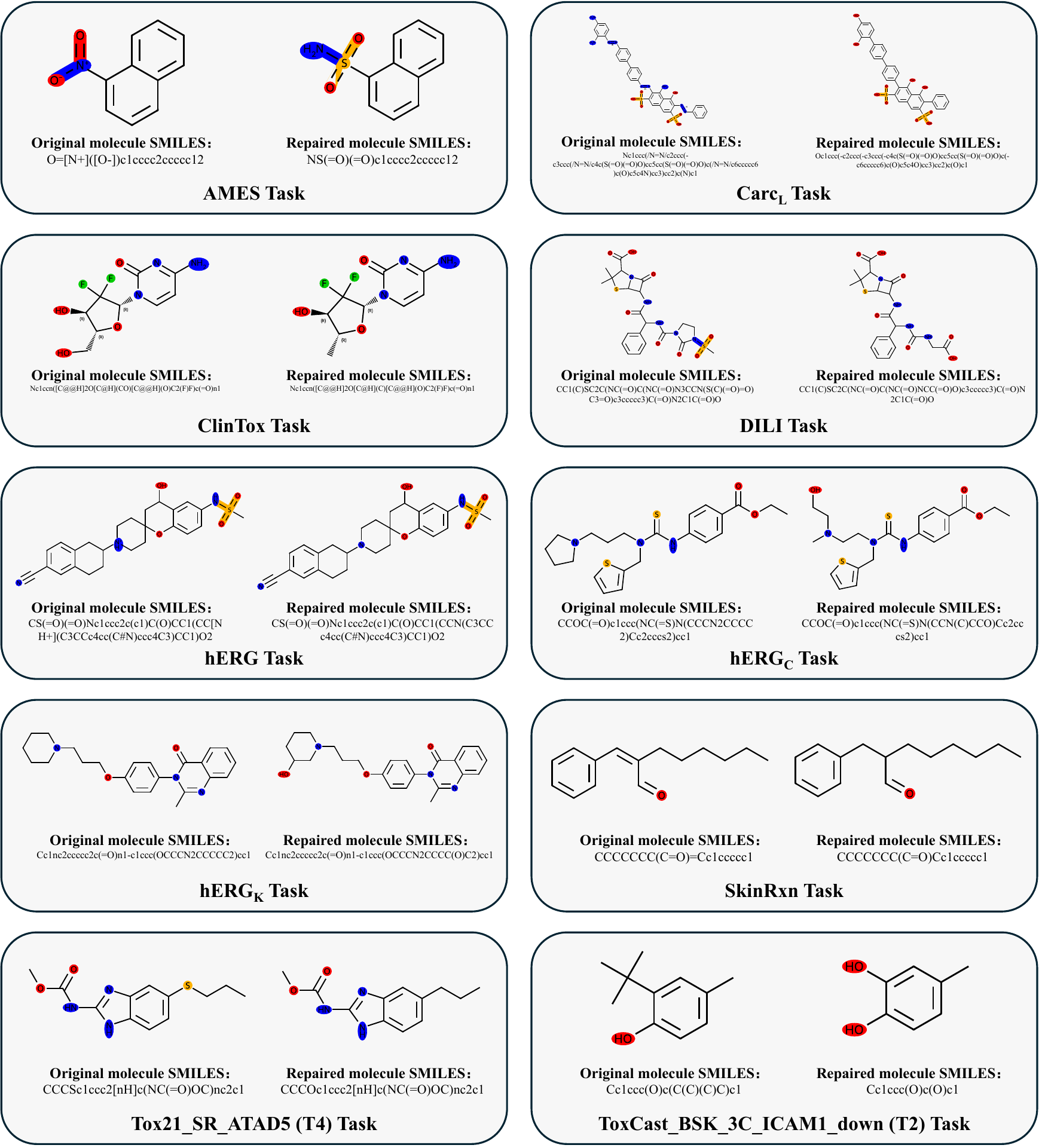}
  \caption{Representative failed toxicity repair cases of \texttt{Claude Opus 4.5}~\cite{anthropic_claude_opus4_5_2025}.}
  \label{fig:failed_cases}
\end{figure}

This section selects the MLLM with the highest repair success rate—\texttt{Claude Opus 4.5}~\cite{anthropic_claude_opus4_5_2025}—and presents its successful and failed toxicity repair cases across tasks~\cite{anthropic_claude_opus4_5_2025}. All selected examples passed the full ToxiEval evaluation chain. Given that this model achieves a success rate of 0 on the \texttt{LD50} task, \texttt{LD50} is excluded from the case analysis in this section. For \texttt{Tox21} and \texttt{ToxCast}, we randomly select one subtask from their respective subtask sets and, within that subtask, randomly choose one successful and one failed sample; for the remaining tasks, we directly randomly sample one successful and one failed instance from the corresponding pools. For the selected molecules, we adopt the same rendering settings as in the input stage and visualize their 2D molecular structures using RDKit. Fig.~\ref{fig:success_cases} shows successful repair cases, and Fig.~\ref{fig:failed_cases} shows failed repair cases.

We conducted manual inspection on the selected successful and failed samples, focusing on structural validity, the absence of clearly unreasonable chemical motifs, and whether the edits are primarily constrained to the original molecular scaffold. No obviously chemically unreasonable structures were observed in the inspected samples. This result supports the effectiveness of the structure validity check in the evaluation chain and further indicates that \texttt{Claude Opus 4.5} demonstrates stable instruction-following behavior under structural similarity constraints (Structural Similarity, $\mathcal{S}_{\text{sim}}$).

\end{document}